\documentclass[10pt,twocolumn,letterpaper]{article}

\usepackage{wacv}              

\definecolor{wacvblue}{rgb}{0.21,0.49,0.74}
\usepackage[pagebackref,breaklinks,colorlinks,allcolors=wacvblue]{hyperref}
\usepackage{booktabs}
\usepackage{multirow}
\usepackage{pifont}
\usepackage{colortbl}
\usepackage{makecell}
\usepackage{xcolor}
\usepackage{subcaption}
\usepackage{amsmath,mathtools}
\usepackage{graphicx}
\usepackage{wrapfig}
\usepackage[accsupp]{axessibility}

\def\wacvPaperID{2317} 
\def\confName{WACV}
\def\confYear{2026}

\title{PatchEAD: Unifying Industrial Visual Prompting Frameworks for Patch-Exclusive Anomaly Detection}

\author{
Po-Han Huang\textsuperscript{1}\quad
Jeng-Lin Li\textsuperscript{1}\quad
Po-Hsuan Huang\textsuperscript{1}\quad
Ming-Ching Chang\textsuperscript{2}\quad
Wei-Chao Chen\textsuperscript{1}\\
\textsuperscript{1}Inventec Corporation \quad
\textsuperscript{2}University at Albany, State University of New York\\
{\tt\small \{huang.po-han, li.johncl, huang.reese, chen.wei-chao\}@inventec.com \quad mchang2@albany.edu}
}

\begin{document}

\maketitle
\begin{abstract}
Industrial anomaly detection is increasingly relying on foundation models, aiming for strong out-of-distribution generalization and rapid adaptation in real-world deployments. Notably, past studies have primarily focused on textual prompt tuning, leaving the intrinsic visual counterpart fragmented into processing steps specific to each foundation model. We aim to address this limitation by proposing a unified patch-focused framework, {\bf Patch-Exclusive Anomaly Detection (PatchEAD)}, enabling training-free anomaly detection that is compatible with diverse foundation models. The framework constructs visual prompting techniques, including an alignment module and foreground masking. Our experiments show superior few-shot and batch zero-shot performance compared to prior work, despite the absence of textual features. 
Our study further examines how backbone structure and pretrained characteristics affect patch-similarity robustness, providing actionable guidance for selecting and configuring foundation models for real-world visual inspection. These results confirm that a well-unified patch-only framework can enable quick,  calibration-light deployment without the need for carefully engineered textual prompts.

\end{abstract}    
\section{Introduction}
\label{sec:intro}




Anomaly detection is a crucial area in computer vision that automates the identification of irregular regions in images, with applications for ensuring product quality and reducing costs in industrial settings. To avoid the need for collecting anomalous data, early research has increasingly focused on unsupervised, few-shot, or zero-shot learning methods. These approaches train models on a large number of normal samples, typically using either reconstruction-based~\cite{DRAEM, UniAD, reconpatch, MambaAD} or feature-based~\cite{trustmae, pni, PaDiM, patchcore} techniques. However, poor generalization to new domains or previously unseen anomalies remains a major challenge for real-world implementation.

Foundation models pre-trained on large-scale datasets have recently demonstrated strong generalizability across domains and high accuracy on downstream tasks~\cite{liu2024few, zhang2023survey}. Building on this, several works have leveraged such models for anomaly detection~\cite{WinCLIP, AdaCLIP, AnomalyCLIP, VCPCLIP, InCTRL}, pushing zero-shot and few-shot performance to new benchmarks. Despite this, the {\em one-model-fit-all} paradigm is not fully realized, as most approaches still rely heavily on additional module training and prompt tuning. The subsequent designs sometimes depend on the properties of the chosen foundation models. The complexity reduces adaptability in fast-evolving industrial environments~\cite{cao2023segment}, highlighting the need to unify a pipeline for foundation model deployments. 

\begin{figure}[t]
\centerline{
   \includegraphics[width=\linewidth]{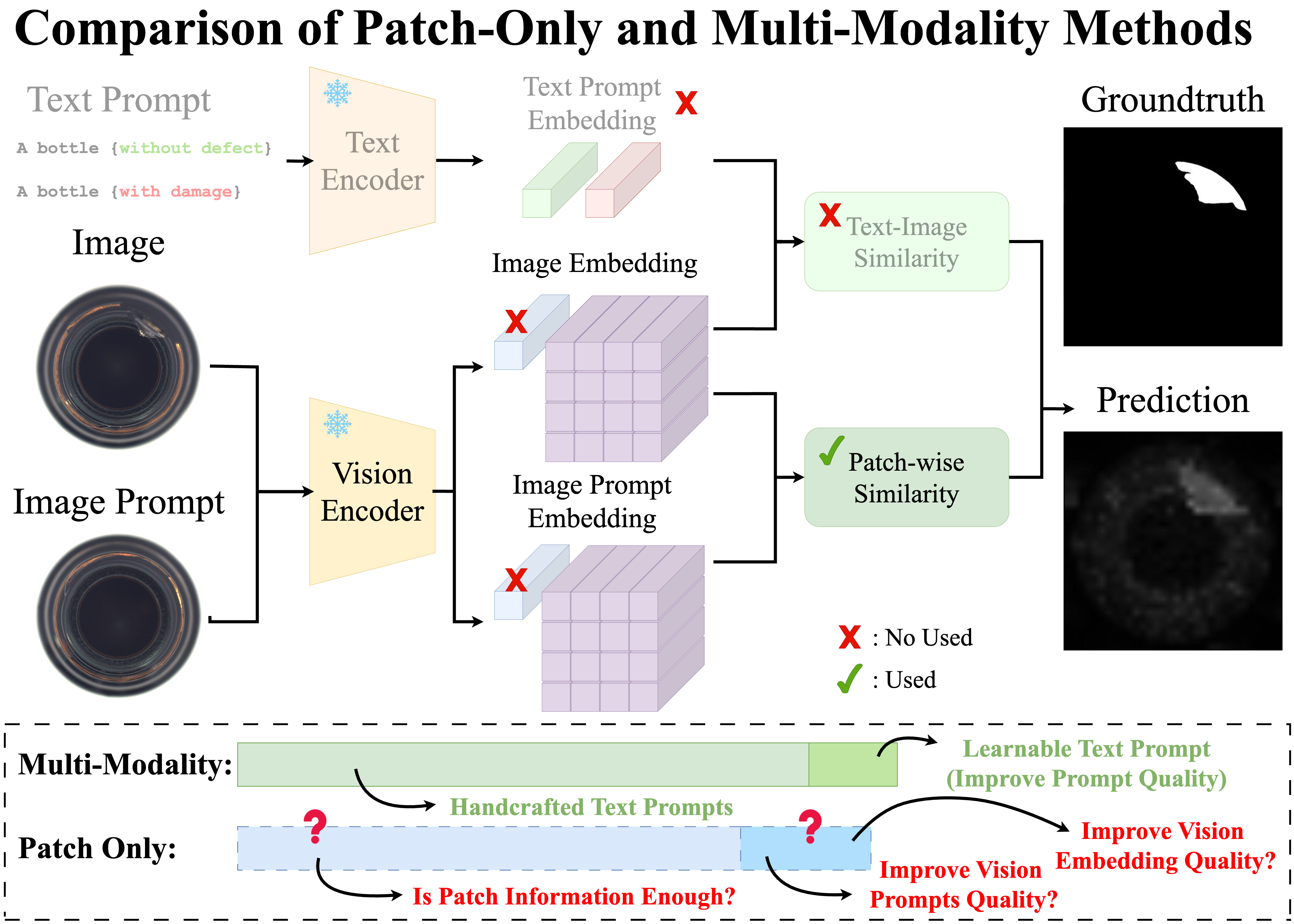}
}
\caption{Comparison of patch-only (PatchEAD) and multi-modality industrial anomaly detection methods. We investigate whether a purely visual approach can achieve competitive performance of multi-modality methods, and whether further optimization of vision prompts can enhance effectiveness.}
\label{fig:teaser}
\end{figure}




Intriguingly, we observe that previous studies mostly shape the pipeline using only textual prompt tuning, despite employing vision-language foundation models. A practical research question arises: {\em how can a unified pipeline be constructed using visual cues to support diverse modern foundation models?} In contrast to textual prompts expressing all evidence through natural language, a pipeline with \textit{patch-level visual prompts} requires a series of preprocessing and calibration designs for normal references. For example, the success of AnomalyDINO~\cite{AnomalyDINO} is largely attributed to its tailored pipeline for DINOv2~\cite{dinov2}, which was pretrained via self-supervised distillation for image representation learning. However, it remains unclear whether the same pipeline can be applied to CLIP or newer Perceptual Encoder (PE) models. The shortcoming of lacking a unified visual procedure parallels the challenge of highly varied textual prompting methods, limiting usability across scenarios. Therefore, we aim to identify a unified patch-only visual pipeline for training-free anomaly detection, as shown in Figure~\ref{fig:teaser}.

\begin{figure*}[t]
\centerline{
  \includegraphics[width=\textwidth]{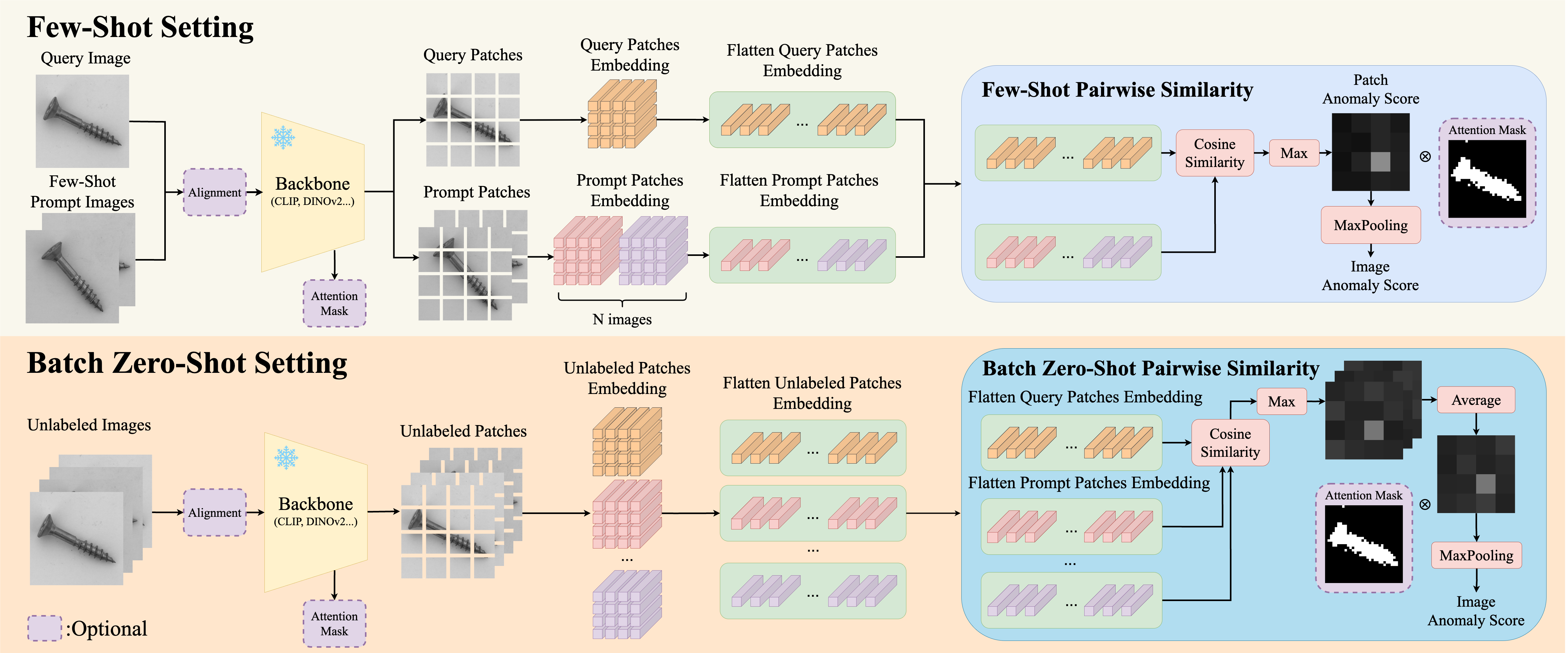}
}
\caption{
{\bf The {\em PatchEAD} framework in few-shot and batch zero-shot settings:} In \textbf{few-shot}, normal prompt and query images are passed through the backbone to extract patch embeddings, which are flattened to compute patch-wise cosine similarities. The highest patch anomaly score is used as the image-level score. In \textbf{batch Zero-shot},each image is compared to the rest of the batch using leave-one-out similarity, with its image-level anomaly score computed as the average of these scores. The version with optional alignment and masking modules is denoted as PatchEAD$^+$.
}
\label{fig:framework}
\end{figure*}

Following these observations, we introduce {\bf Patch-Exclusive Anomaly Detection (PatchEAD)}, a framework based on patch-based feature comparison that excludes multi-modal features such as text and global image features. PatchEAD ensures high compatibility with any pre-trained image-based model while maintaining anomaly detection performance. PatchEAD is designed for both batch zero-shot and few-shot settings (Figure~\ref{fig:framework}), requiring no additional training, with a backbone-agnostic, vision-only patch-similarity formulation that standardizes usage across heterogeneous backbones and datasets. This design enhances industrial applicability and offers flexibility for future extension to different backbones.

We comprehensively evaluated PatchEAD on seven industrial anomaly datasets, comparing its performance with seven state-of-the-art (SoTA) few-shot~\cite{AprilGAN, AnomalyGPT, PromptAD, InCTRL, PaDiM, patchcore, WinCLIP} and four zero-shot or batch zero-shot~\cite{AnomalyCLIP, AdaCLIP, WinCLIP, MuSc} anomaly detection methods. It achieves 96\% image-level AUC on MVTec and 89.5\% on VisA in the few-shot setting, and 95.3\% on MVTec and 87.6\% on VisA in the batch zero-shot setting. These results highlight its strong performance, on par with SoTA methods that require training or additional modalities.

Our contribution is summarized as follows:

\begin{itemize}[leftmargin=10pt] \itemsep -.1em
\item We investigate a practical foundation model problem in anomaly detection: the lack of unified visual-prompting pipeline. Our experiments address the generalization limitations of previous works that tailor visual processing to specific foundation models. 

\item We present the PatchEAD framework, characterized by its patch-only design, training-free nature, and compatibility with diverse foundation models. PatchEAD achieves SoTA anomaly detection results, underscoring the effectiveness of the unified visual pipeline even without relying on textual features.

\item We further integrate alignment and background removal to create an enhanced version, PatchEAD$^+$, which improves generalization and performance. This approach demonstrates the potential of a vision-centric augmentation strategy to strengthen patch-based frameworks.
\end{itemize}


\section{Related Work}
\label{sec:relatedwork}


\subsection{Unsupervised Anomaly Detection}

The limited availability of abnormal samples often restricts current anomaly detection methods, making unsupervised learning approaches essential. One-class classification methods~\cite{cutpaste, memseg, simplenet, generalad} detect anomalies by modeling the distribution of normal samples and flagging outliers as anomalies. 
Reconstruction-based methods~\cite{DRAEM, UniAD, reconpatch, MambaAD} 
train autoencoders on large sets of normal samples, ensuring that abnormal samples are poorly reconstructed, thereby revealing anomaly regions. Knowledge distillation methods~\cite{multiresolution, remembering} use discrepancies between teacher and student networks to identify anomalies. Memory bank methods~\cite{trustmae, pni, PaDiM} store features of normal images and detect anomalies by measuring feature distances. 
PatchCore~\cite{patchcore} in particular demonstrates the effectiveness of local patch features, which outperform global image features in trained memory banks. Despite promising results, these methods often struggle to generalize across datasets, as they are constrained by training data size and feature extractor capacity.




\subsection{Foundation Models in Anomaly Detection}

Foundation models are widely used in recent anomaly detection methods due to their broad generalization capabilities. AnomalyCLIP~\cite{AnomalyCLIP} learns generalizable object-agonistic text prompts to adapt CLIP for distinguishing normal and abnormal cases. VCP-CLIP~\cite{VCPCLIP} uses visual context prompts to enhance CLIP's ability to detect anomaly semantics, while PromptAD~\cite{PromptAD} appends abnormal suffixes to normal prompts and introduces an explicit anomaly margin for prompt learning. InCTRL~\cite{InCTRL} identifies anomalies by learning residuals between query images and normal prompts, improving cross-application generalization. AdaCLIP~\cite{AdaCLIP} incorporates learnable hybrid prompts for better adaptability in CLIP. These methods, however, often add significant computational overhead through prompt learning or additional model training.

Training-free approaches~\cite{WinCLIP, AnoVL} generally use handcrafted text prompts, as in WinCLIP~\cite{WinCLIP}, which aligns multi-scale image features with a manually designed text prompt. Recently, MuSc~\cite{MuSc} computes anomaly scores by comparing unlabeled test image patches, while AnomalyDINO~\cite{AnomalyDINO} stores nominal patch representations from the reference samples in a memory bank and compares test patch features with these nominal representations. 

Among these studies, only WinCLIP~\cite{WinCLIP} explicitly evaluates applicability to both few-shot and zero-shot settings. Additionally, MuSc and AnomalyDINO adopt relatively complex multi-module designs, such as multi-scale aggregation or PCA-based mask extraction, which pose challenges for practical deployment.
In contrast, our work emphasizes the differences in patch information, offering a comprehensive investigation of few-shot and batch zero-shot settings, demonstrating the effectiveness of a training-free, patch-only approach.





\section{Method}
\vspace{-1mm}

Inspired by PatchCore~\cite{patchcore}, which shows that aggregating local patches preserves sufficient spatial information and avoids biases from the pretraining space, we first introduce our framework with few-shot and batch zero-shot settings in $\S$\ref{ssec:PatchEAD}. $\S$\ref{ssec:Cross:Patch:Scoring} details the derivation of anomaly scores, and $\S$\ref{ssec:robust} presents techniques to improve the robustness of our patch-wise framework.

\begin{table*}[t]
\caption{
{\bf Comparisons of few-shot settings with state-of-the-art methods} on MVTec and VisA datasets. Scores are reported as $mean_{\pm std}$; training-based methods are shown in \textcolor{gray}{gray}, and the highest score is highlighted in \textcolor{red}{\bf bold red}.
}
\centerline{
\setlength{\tabcolsep}{1.5mm}
\resizebox{\textwidth}{!}{
\begin{tabular}{c|c|c|c|c|c|c|c|c|c|c|c}
    \toprule
    \multirow{3}{*}{Settings} & \multirow{3}{*}{Methods} & \multirow{3}{*}{\makecell{Training\\Free}} & \multirow{3}{*}{\makecell{Patch\\Only}} & \multicolumn{4}{c|}{\textbf{MVTec}} & \multicolumn{4}{c}{\textbf{VisA}} \\ \cline{5-12}
    & & & & \multicolumn{2}{|c|}{Image-Level} & \multicolumn{2}{c|}{Pixel-Level} & \multicolumn{2}{c|}{Image-Level} & \multicolumn{2}{c}{Pixel-Level} \\ \cline{5-12}
    & & & & AUC & AP & AUC & PRO & AUC & AP & AUC & PRO \\
    \hline\hline 
    \multirow{11}{*}{\centering 1-shot} & 
    \textcolor{gray}{APIRL-GAN~\cite{AprilGAN}} &  &  & 
    \textcolor{gray}{$92.0_{\pm 0.3}$} & 
    \textcolor{gray}{$95.8_{\pm 0.2}$} & 
    \textcolor{gray}{$95.1_{\pm 0.1}$} & 
    \textcolor{gray}{$90.6_{\pm 0.2}$} & 
    \textcolor{gray}{$91.2_{\pm 0.8}$} & 
    \textcolor{gray}{$93.3_{\pm 0.8}$} & 
    \textcolor{gray}{$96.0_{\pm 0.0}$} & 
    \textcolor{gray}{$90.0_{\pm 0.1}$} \\
    & 
    \textcolor{gray}{AnomalyGPT~\cite{AnomalyGPT}} &  &  & 
    \textcolor{gray}{$94.1_{\pm 1.1}$} & 
    \textcolor{gray}{-} & 
    \textcolor{gray}{$95.3_{\pm 0.1}$} & 
    \textcolor{gray}{-} & 
    \textcolor{gray}{$87.4_{\pm 0.8}$} & 
    \textcolor{gray}{-} & 
    \textcolor{gray}{$96.2_{\pm 0.1}$} & 
    \textcolor{gray}{-} \\
    & 
    \textcolor{gray}{PromptAD~\cite{PromptAD}} &  &  & 
    \textcolor{gray}{$94.6_{\pm 1.7}$} & 
    \textcolor{gray}{-} & 
    \textcolor{gray}{$95.9_{\pm 0.5}$} & 
    \textcolor{gray}{-} & 
    \textcolor{gray}{$86.9_{\pm 2.3}$} & 
    \textcolor{gray}{-} & 
    \textcolor{gray}{$96.7_{\pm 0.4}$} & 
    \textcolor{gray}{-} \\
    & 
    \textcolor{gray}{InCTRL~\cite{InCTRL}} &   &  & 
    \textcolor{gray}{-} & 
    \textcolor{gray}{-} & 
    \textcolor{gray}{-} & 
    \textcolor{gray}{-} & 
    \textcolor{gray}{-} &  
    \textcolor{gray}{-} & 
    \textcolor{gray}{-} & 
    \textcolor{gray}{-} \\
    \cline{2-12} 
    &
    PaDiM~\cite{PaDiM} & \ding{51} & \ding{51} & 
    $76.6_{\pm 3.1}$ & 
    $88.1_{\pm 1.7}$ & 
    $89.3_{\pm 0.9}$ & 
    $73.3_{\pm 2.0}$ & 
    $62.8_{\pm 5.4}$ & 
    $68.3_{\pm 4.0}$ & 
    $89.9_{\pm 0.8}$ & 
    $64.3_{\pm 2.4}$ \\
    &
    PatchCore~\cite{patchcore} & \ding{51} & \ding{51} & 
    $83.4_{\pm 3.0}$ & 
    $92.2_{\pm 1.5}$ & 
    $92.0_{\pm 1.0}$ & 
    $79.7_{\pm 2.0}$ & 
    $79.9_{\pm 2.9}$ & 
    $82.8_{\pm 2.3}$ & 
    $95.4_{\pm 0.6}$ & 
    $80.5_{\pm 2.5}$ \\
    & WinCLIP~\cite{WinCLIP} & \ding{51}  &  & ${93.1_{\pm 2.0}}$ & $\bf 96.5_{\pm 0.9}$ & ${95.2_{\pm 0.5}}$ & $87.1_{\pm 1.2}$ & ${83.8_{\pm 4.0}}$ & ${85.1_{\pm 4.0}}$ & ${96.4_{\pm 0.4}}$ & ${85.1_{\pm 2.1}}$ \\
    & PatchEAD (CLIP) & \ding{51}  & \ding{51} & $92.8_{\pm 0.9}$ & $95.4_{\pm 1.0}$ & $94.6_{\pm 0.2}$ & $87.4_{\pm 0.4}$ & $84.5_{\pm 1.3}$ & $86.6_{\pm 1.1}$ & $94.5_{\pm 0.2}$ & $75.7_{\pm 0.8}$ \\
    & PatchEAD$^+$(CLIP) & \ding{51}  & \ding{51} & $93.4_{\pm 0.8}$ & $96.2_{\pm 0.6}$ & $94.9_{\pm 0.2}$ & $88.0_{\pm 0.5}$ & $ 86.2_{\pm 0.6}$ & $87.7_{\pm 0.9}$ & $95.6_{\pm 0.1}$ & $79.8_{\pm 0.1}$ \\
    & PatchEAD (DINOv2) & \ding{51}  & \ding{51} & $93.6_{\pm 0.7}$ & $95.9_{\pm 0.2}$ & $96.0_{\pm 0.1}$ & {$91.5_{\pm 0.1}$} & $85.4_{\pm 0.3}$ & $85.9_{\pm 0.3}$ & $97.3_{\pm 0.1}$ & $89.8_{\pm 0.2}$ \\
    & PatchEAD$^+$(DINOv2) & \ding{51}  & \ding{51} & $\textcolor{red}{\bf 95.0_{\pm 0.6}}$ & $\textcolor{red}{\bf97.0_{\pm 0.2}}$ & $\textcolor{red}{\bf 96.1_{\pm 0.2}}$ & $\textcolor{red}{\bf 91.6_{\pm 0.1}}$ & $\textcolor{red}{\bf89.1_{\pm 0.4}}$ & $\textcolor{red}{\bf 89.7_{\pm 0.5}}$ & $\textcolor{red}{\bf 97.4_{\pm 0.1}}$ & $\textcolor{red}{\bf 90.6_{\pm 0.2}}$ \\
    \hline\hline
    \multirow{11}{*}{\centering 2-shot} & 
    \textcolor{gray}{APIRL-GAN~\cite{AprilGAN}} &  &  & \textcolor{gray}{$92.4_{\pm 0.3}$} & 
    \textcolor{gray}{$96.0_{\pm 0.2}$} & 
    \textcolor{gray}{$95.5_{\pm 0.0}$} & 
    \textcolor{gray}{$91.3_{\pm 0.1}$} & 
    \textcolor{gray}{$92.2_{\pm 0.3}$} & 
    \textcolor{gray}{$94.2_{\pm 0.3}$} &
    \textcolor{gray}{$96.2_{\pm 0.0}$} &
    \textcolor{gray}{$90.1_{\pm 0.1}$} \\
    & \textcolor{gray}{AnomalyGPT~\cite{AnomalyGPT}} &  &  & 
    \textcolor{gray}{$95.5_{\pm 0.8}$} & 
    \textcolor{gray}{-} & 
    \textcolor{gray}{$95.6_{\pm 0.2}$} & 
    \textcolor{gray}{-} & 
    \textcolor{gray}{$88.6_{\pm 0.7}$} & 
    \textcolor{gray}{-} & 
    \textcolor{gray}{$96.4_{\pm 0.1}$} & 
    \textcolor{gray}{-} \\
    & \textcolor{gray}{PromptAD~\cite{PromptAD}} &  &  & 
    \textcolor{gray}{$95.7_{\pm 1.5}$} & 
    \textcolor{gray}{-} & 
    \textcolor{gray}{$96.2_{\pm 0.3}$} & 
    \textcolor{gray}{-} & 
    \textcolor{gray}{$88.3_{\pm 2.0}$} & 
    \textcolor{gray}{-} & 
    \textcolor{gray}{$97.1_{\pm 0.3}$} & 
    \textcolor{gray}{-} \\
    & 
    \textcolor{gray}{InCTRL~\cite{InCTRL}} &  &  & 
    \textcolor{gray}{$94.0_{\pm 1.5}$} & 
    \textcolor{gray}{$96.9_{\pm 0.4}$} & 
    \textcolor{gray}{-} & 
    \textcolor{gray}{-} & 
    \textcolor{gray}{$85.8_{\pm 2.2}$} & 
    \textcolor{gray}{$87.7_{\pm 1.6}$} & 
    \textcolor{gray}{-} & 
    \textcolor{gray}{-} \\
    \cline{2-12}
    &
    PaDiM~\cite{PaDiM} & \ding{51} & \ding{51} & 
    $78.9_{\pm 3.1}$ & 
    $89.3_{\pm 1.7}$ & 
    $91.3_{\pm 0.7}$ & 
    $78.2_{\pm 1.8}$ & 
    $67.4_{\pm 5.1}$ & 
    $71.6_{\pm 3.8}$ & 
    $92.0_{\pm 0.7}$ & 
    $70.1_{\pm 2.6}$ \\
    &
    PatchCore~\cite{patchcore} & \ding{51} & \ding{51} & 
    $86.3_{\pm 3.3}$ & 
    $93.8_{\pm 1.7}$ & 
    $93.3_{\pm 0.6}$ & 
    $82.3_{\pm 1.3}$ & 
    $81.6_{\pm 4.0}$ & 
    $84.8_{\pm 3.2}$ & 
    $96.1_{\pm 0.5}$ & 
    $82.6_{\pm 2.3}$ \\
    & WinCLIP~\cite{WinCLIP} & \ding{51}  &  & ${94.4_{\pm 1.3}}$ & ${97.0_{\pm 0.7}}$ & ${96.0_{\pm 0.3}}$ & $88.4_{\pm 0.9}$ & $84.6_{\pm 2.4}$ & $85.8_{\pm 2.7}$ & ${96.8_{\pm 0.3}}$ & ${86.2_{\pm 1.4}}$ \\
    & PatchEAD (CLIP) & \ding{51}  & \ding{51} & $92.8_{\pm 0.5}$ & $95.5_{\pm 0.4}$ & $94.8_{\pm 0.1}$ & $87.6_{\pm 0.3}$ & $87.8_{\pm 1.4}$ & {$89.4_{\pm 1.7}$} & $95.6_{\pm 0.1}$ & $78.8_{\pm 0.2}$ \\
    & PatchEAD$^+$(CLIP) & \ding{51}  & \ding{51} & $93.1_{\pm 0.4}$ & $95.9_{\pm 0.4}$ & $95.2_{\pm 0.2}$ & $88.6_{\pm 0.2}$ & $87.6_{\pm 1.2}$ & $88.8_{\pm 1.7}$ & $96.2_{\pm 0.1}$ & $82.0_{\pm 1.0}$ \\
    & PatchEAD (DINOv2) & \ding{51}  & \ding{51} & ${95.0_{\pm 0.1}}$ & $96.6_{\pm 0.3}$ & $96.3_{\pm 0.1}$ & ${92.1_{\pm 0.1}}$ & ${87.7_{\pm 0.3}}$ & $88.0_{\pm 0.5}$ & $\textcolor{red}{\bf 97.7_{\pm 0.1}}$ & $90.8_{\pm 0.2}$ \\
    & PatchEAD$^+$(DINOv2) & \ding{51}  & \ding{51} & $\textcolor{red}{\bf 96.2_{\pm 0.4}}$ & $\textcolor{red}{\bf 97.6_{\pm 0.5}}$ & $\textcolor{red}{\bf 96.4_{\pm 0.1}}$ & $\textcolor{red}{\bf 92.2_{\pm 0.1}}$ & $\textcolor{red}{\bf 90.6_{\pm 0.5}}$ & $\textcolor{red}{\bf 91.0_{\pm 0.4}}$ & $\textcolor{red}{\bf 97.7_{\pm 0.1}}$ & $\textcolor{red}{\bf 91.3_{\pm 0.1}}$ \\
    \hline\hline
    \multirow{11}{*}{\centering 4-shot} & 
    \textcolor{gray}{APIRL-GAN~\cite{AprilGAN}} &  &  & \textcolor{gray}{$92.8_{\pm 0.2}$} & 
    \textcolor{gray}{$96.3_{\pm 0.1}$} & 
    \textcolor{gray}{$95.9_{\pm 0.0}$} & 
    \textcolor{gray}{$91.8_{\pm 0.1}$} & 
    \textcolor{gray}{$92.6_{\pm 0.4}$} & 
    \textcolor{gray}{$94.5_{\pm 0.3}$} & 
    \textcolor{gray}{$96.2_{\pm 0.0}$} & 
    \textcolor{gray}{$90.2_{\pm 0.1}$} \\
    & \textcolor{gray}{AnomalyGPT~\cite{AnomalyGPT}} &   &  & \textcolor{gray}{$96.3_{\pm 0.3}$} & \textcolor{gray}{-} & \textcolor{gray}{$96.2_{\pm 0.1}$} & \textcolor{gray}- & \textcolor{gray}{$90.6_{\pm 0.7}$} & \textcolor{gray}- & \textcolor{gray}{$96.7_{\pm 0.1}$} & \textcolor{gray}- \\
    & \textcolor{gray}{PromptAD~\cite{PromptAD}} &   &  & \textcolor{gray}{$96.6_{\pm 0.9}$} & \textcolor{gray}- & \textcolor{gray}{$96.5_{\pm 0.2}$} & \textcolor{gray}- & \textcolor{gray}{$89.1_{\pm 1.7}$} & \textcolor{gray}- & \textcolor{gray}{$97.4_{\pm 0.3}$} & \textcolor{gray}- \\
    & \textcolor{gray}{InCTRL~\cite{InCTRL}} &   &  & \textcolor{gray}{$94.5_{\pm 1.8}$} & \textcolor{gray}{$97.2_{\pm 0.6}$} & \textcolor{gray}{-} & \textcolor{gray}{-} & \textcolor{gray}{$87.7_{\pm 1.9}$} & \textcolor{gray}{$90.2_{\pm 2.7}$} & \textcolor{gray}{-} & \textcolor{gray}{-} \\
    \cline{2-12}
    &
    PaDiM~\cite{PaDiM} & \ding{51} & \ding{51} & 
    $80.4_{\pm 2.5}$ & 
    $90.5_{\pm 1.6}$ & 
    $92.6_{\pm 0.7}$ & 
    $81.3_{\pm 1.9}$ & 
    $72.8_{\pm 2.9}$ & 
    $75.6_{\pm 2.2}$ & 
    $93.2_{\pm 0.5}$ & 
    $72.6_{\pm 1.9}$ \\
    &
    PatchCore~\cite{patchcore} & \ding{51} & \ding{51} & 
    $88.8_{\pm 2.6}$ & 
    $94.5_{\pm 1.5}$ & 
    $94.3_{\pm 0.5}$ & 
    $84.3_{\pm 1.6}$ & 
    $85.3_{\pm 2.1}$ & 
    $87.5_{\pm 2.1}$ & 
    $96.8_{\pm 0.3}$ & 
    $84.9_{\pm 1.4}$ \\
    & WinCLIP~\cite{WinCLIP} & \ding{51}  &  & $95.2_{\pm 1.3}$ & $97.3_{\pm 0.6}$ & ${96.2_{\pm 0.3}}$ & $89.0_{\pm 0.8}$ & $87.3_{\pm 1.8}$ & ${88.8_{\pm 1.8}}$ & ${97.2_{\pm 0.2}}$ & ${87.6_{\pm 0.9}}$ \\
    & PatchEAD (CLIP) & \ding{51}  & \ding{51} & $95.2_{\pm 0.9}$ & $97.2_{\pm 0.5}$ & $95.6_{\pm 0.1}$ & $88.9_{\pm 0.4}$ & $89.3_{\pm 0.5}$ & $90.2_{\pm 0.9}$ & $96.1_{\pm 0.1}$ & $80.4_{\pm 0.1}$ \\
    & PatchEAD$^+$(CLIP) & \ding{51}  & \ding{51} & $95.2_{\pm 0.8}$ & $97.2_{\pm 0.5}$ & $95.9_{\pm 0.1}$ & $89.6_{\pm 0.3}$ & $89.5_{\pm 0.3}$ & $90.5_{\pm 0.3}$ & $96.7_{\pm 0.1}$ & $82.8_{\pm 0.5}$ \\
    & PatchEAD (DINOv2) & \ding{51}  & \ding{51} & $96.0_{\pm 1.2}$ & $97.8_{\pm 0.7}$ & ${96.8_{\pm 0.1}}$ & ${92.8_{\pm 0.1}}$ & ${88.9_{\pm 0.1}}$ & $88.9_{\pm 0.3}$ & $\textcolor{red}{\bf 98.0_{\pm 0.1}}$ & $91.5_{\pm 0.2}$ \\
    & PatchEAD$^+$(DINOv2) & \ding{51}  & \ding{51} & $\textcolor{red}{\bf 97.0_{\pm 0.8}}$ & $\textcolor{red}{\bf 98.3_{\pm 0.4}}$ & $\textcolor{red}{\bf 96.9_{\pm 0.1}}$ & $\textcolor{red}{\bf 92.9_{\pm 0.2}}$ & $\textcolor{red}{\bf 91.9_{\pm 0.3}}$ & $\textcolor{red}{\bf 92.3_{\pm 0.6}}$ & $\textcolor{red}{\bf 98.0_{\pm 0.1}}$ & $\textcolor{red}{\bf 92.1_{\pm 0.1}}$ \\
    \bottomrule
\end{tabular}
}
}
\label{tab:few_shot_comparison}
\end{table*}
\subsection{Patch-Exclusive Anomaly Detection}
\label{ssec:PatchEAD}


We emphasize the pivotal role of patch features and introduce the Patch-Exclusive Anomaly Detection (PatchEAD) framework (Figure~\ref{fig:framework}). PatchEAD combines patch feature extraction with cross-patch scoring modules and supports both few-shot and batch zero-shot anomaly detection. By relying solely on patch-level information, it achieves robust, generalized performance.

Each whole image input $X\in\mathbb{R}^{W\times H}$, with width $W$ and height $H$, is divided into equally sized, non-overlapping patches, forming a grid across the entire image. These patches are then individually processed by a frozen, pre-trained vision encoder $f_v(\cdot)$, which extracts feature embeddings for each patch. This method ensures that each patch independently represents a localized area of the image, capturing fine-grained spatial details while minimizing redundancy across patches. This results in patch-level token embedding maps $E\in \mathbb{R}^{L\times D}$, where $L$ denotes the number of patches and $D$ is the embedding dimension, representing the visual characteristics of each patch. 


We adopt a flexible choice of pre-trained vision encoder $f_v(\cdot)$—including, but not limited to, models trained with self-supervised teacher-student objectives ({\em e.g.}, DINOv2~\cite{dinov2}), cross-modality retrieval objectives ({\em e.g.}, CLIP~\cite{CLIP2021}), or any of the backbones evaluated in our ablation study. In each case, we retain only the visual branch---for example, dropping the [CLS] token in DINOv2 and removing the text encoder $f_t$ in CLIP, ensuring that all patch features come purely from visual representations.

In the \textbf{few-shot} setting, we use $N$ normal prompt images to establish anchors that enable the model to distinguish anomalies. In the \textbf{batch zero-shot} setting, framed here as an unsupervised anomaly-detection scenario, no precollected normal library is required; instead, we operate on a batch of unlabeled query images under the assumption that most samples in the test batch are normal. This differs from traditional methods that rely on a fixed pool of normal references. Specifically, following ACR~\cite{ACR} and MuSc~\cite{MuSc}, we treat each image in the batch as a query and use the remaining images as on-the-fly normal references to generate an anomaly prediction for that query. Repeating this process for every image yields a complete set of anomaly scores.


\begin{table*}[t]
\caption{{\bf Comparisons of batch zero-shot {\em image-level} (top) and {\em pixel-level} (bottom) performance with state-of-the-art methods} across multiple datasets. 
Training-based methods are shown in \textcolor{gray}{gray}, and the highest score is highlighted in \textcolor{red}{bold red}. Base versions of all networks are used for fair comparison at similar model sizes. $^\dagger$ indicates methods reproduced using official code and weights.
}
\centerline{
\resizebox{\textwidth}{!}{
\setlength{\tabcolsep}{1.5mm}
\begin{tabular}{c|c|c|c|c|c|c|c|c|c|c|c|c|c|c|c|c|c|c}
    \toprule
    \multirow{2}{*}{{\bf Image-level evaluation}} & \multirow{2}{*}{\makecell{Training\\Free}} & \multirow{2}{*}{\makecell{Patch\\Only}} & \multicolumn{2}{c|}{MVTec\cite{mvtec}} & \multicolumn{2}{c|}{VisA\cite{visa}} & \multicolumn{2}{c|}{MPDD\cite{MPDD}} & \multicolumn{2}{c|}{BTAD\cite{BTAD}} & \multicolumn{2}{c|}{KSDD\cite{KSDD}} & \multicolumn{2}{c|}{DAGM\cite{DAGM}} & \multicolumn{2}{c|}{DTD-Syn.\cite{DTD-Synthetic}} & \multicolumn{2}{c}{Average} \\ \cline{4-19}
    & & & AUC & AP & AUC & AP & AUC & AP & AUC & AP & AUC & AP & AUC & AP & AUC & AP & AUC & AP \\
    \hline\hline
    \textcolor{gray}{AnomalyCLIP~\cite{AnomalyCLIP}} & & & 
    \textcolor{gray}{$91.5$} & 
    \textcolor{gray}{${96.2}$} & 
    \textcolor{gray}{$82.1$} & 
    \textcolor{gray}{$85.4$} & 
    \textcolor{gray}{$77.0$} & 
    \textcolor{gray}{$82.0$} & 
    \textcolor{gray}{$88.3$} & 
    \textcolor{gray}{$87.3$} & 
    \textcolor{gray}{$84.7$} & 
    \textcolor{gray}{$80.0$} & 
    \textcolor{gray}{$97.5$} & 
    \textcolor{gray}{$92.3$} & 
    \textcolor{gray}{$93.5$} & 
    \textcolor{gray}{$97.0$} &
    \textcolor{gray}{$87.8$} & 
    \textcolor{gray}{$88.6$} \\
    \textcolor{gray}{AdaCLIP$^\dagger$~\cite{AdaCLIP}} & & & 
    \textcolor{gray}{$89.9$} & 
    \textcolor{gray}{$95.7$} & 
    \textcolor{gray}{$86.1$} & 
    \textcolor{gray}{$88.2$} & 
    \textcolor{gray}{$69.5$} & 
    \textcolor{gray}{$74.6$} & 
    \textcolor{gray}{$90.4$} & 
    \textcolor{gray}{$94.1$} & 
    \textcolor{gray}{$96.9$} & 
    \textcolor{gray}{$91.1$} & 
    \textcolor{gray}{$96.9$} & 
    \textcolor{gray}{$88.2$} & 
    \textcolor{gray}{$91.5$} & 
    \textcolor{gray}{$95.4$} & 
    \textcolor{gray}{$88.7$} & 
    \textcolor{gray}{$89.6$} \\
    \cline{1-19}
    WinCLIP~\cite{WinCLIP} & \ding{51} & & $91.8$ & $96.5$ & $78.1$ & $81.2$ & $63.6$ & $69.9$ & $68.2$ & $70.9$ & $84.3$ & $77.4$ & $91.8$ & $79.5$ & $93.2$ & $92.6$ & $81.6$ & $81.1$ \\
    MuSc$^\dagger$~\cite{MuSc} & \ding{51} &  & $95.8$ & $\textcolor{red}{\bf 98.2}$ & $88.1$ & $90.1$ & $68.6$ & $74.1$ & $\textcolor{red}{\bf 95.4}$ & $\textcolor{red}{\bf 98.9}$ & $92.6$ & $67.6$ & ${95.2}$ & $85.8$ & $90.6$ & $95.8$ & $89.5$ & $87.2$ \\
    PatchEAD (CLIP) & \ding{51} & \ding{51} & $94.5$ & $97.0$ & $86.8$ & $88.0$ & $69.5$ & $75.5$ & $92.3$ & ${96.8}$ & $89.0$ & $63.5$ & $92.5$ & $78.7$ & $92.4$ & $97.2$ & $88.1$ & $85.2$ \\
    PatchEAD$^+$ (CLIP) & \ding{51} & \ding{51} & $94.6$ & $97.1$ & $86.9$ & $88.9$ & $\textcolor{red}{\bf 70.6}$ & $\textcolor{red}{\bf 75.9}$ & $91.5$ & $95.3$ & $89.4$ & $66.3$ & $94.3$ & $80.4$ & $91.9$ & $96.4$ & $88.5$ & $85.8$ \\
    PatchEAD (DINOv2) & \ding{51} & \ding{51} & $94.3$ & $97.0$ & $87.5$ & $87.3$ & $65.3$ & $70.3$ & $92.3$ & $96.0$ & $93.7$ & $67.2$ & $97.6$ & $89.7$ & $91.4$ & ${96.0}$ & $88.9$ & $86.2$ \\
    PatchEAD$^+$ (DINOv2) & \ding{51} & \ding{51} & $\textcolor{red}{\bf 95.9}$ & ${97.9}$ & $\textcolor{red}{\bf 92.2}$ & $\textcolor{red}{\bf 92.6}$ & $68.6$ & ${71.8}$ & ${91.2}$ & $94.3$ & $\textcolor{red}{\bf 97.0}$ & $\textcolor{red}{\bf 85.5}$ & $\textcolor{red}{\bf 98.3}$ & $\textcolor{red}{\bf 92.8}$ & $\textcolor{red}{\bf 97.6}$ & $\textcolor{red}{\bf 98.8}$ & $\textcolor{red}{\bf 91.5}$ & $\textcolor{red}{\bf 90.5}$ \\
    \midrule
\end{tabular}
}
}
\centerline{
\resizebox{\textwidth}{!}{
\setlength{\tabcolsep}{1.5mm}
\begin{tabular}{c|c|c|c|c|c|c|c|c|c|c|c|c|c|c|c|c|c|c}
\toprule
\multirow{2}{*}{{\bf Pixel-level evaluation}} & \multirow{2}{*}{\makecell{Training\\Free}} & \multirow{2}{*}{\makecell{Patch\\Only}} & \multicolumn{2}{c|}{MVTec\cite{mvtec}} & \multicolumn{2}{c|}{VisA\cite{visa}} & \multicolumn{2}{c|}{MPDD\cite{MPDD}} & \multicolumn{2}{c|}{BTAD\cite{BTAD}} & \multicolumn{2}{c|}{KSDD\cite{KSDD}} & \multicolumn{2}{c|}{DAGM\cite{DAGM}} & \multicolumn{2}{c|}{DTD-Syn.\cite{DTD-Synthetic}} & \multicolumn{2}{c}{Average} \\ \cline{4-19}
& & & AUC & PRO & AUC & PRO & AUC & PRO & AUC & PRO & AUC & PRO & AUC & PRO & AUC & PRO & AUC & PRO \\
\hline\hline
    \textcolor{gray}{AnomalyCLIP~\cite{AnomalyCLIP}} &  &  & 
    \textcolor{gray}{${91.1}$} & 
    \textcolor{gray}{${81.4}$} & 
    \textcolor{gray}{$95.4$} & 
    \textcolor{gray}{${87.0}$} & 
    \textcolor{gray}{${96.5}$} & 
    \textcolor{gray}{${88.7}$} & 
    \textcolor{gray}{$94.2$} & 
    \textcolor{gray}{${74.8}$} & 
    \textcolor{gray}{$90.6$} & 
    \textcolor{gray}{${67.8}$} & 
    \textcolor{gray}{$95.6$} & 
    \textcolor{gray}{${91.0}$} & 
    \textcolor{gray}{${97.9}$} & 
    \textcolor{gray}{${92.3}$} & 
    \textcolor{gray}{${94.5}$} & 
    \textcolor{gray}{${83.3}$} \\
    \textcolor{gray}{AdaCLIP$\dagger$~\cite{AdaCLIP}} &  &  & 
    \textcolor{gray}{$89.9$} & 
    \textcolor{gray}{$44.1$} & 
    \textcolor{gray}{${95.8}$} & 
    \textcolor{gray}{$53.5$} & 
    \textcolor{gray}{$92.9$} & 
    \textcolor{gray}{$30.1$} & 
    \textcolor{gray}{${93.7}$} & 
    \textcolor{gray}{$20.3$} & 
    \textcolor{gray}{${95.9}$} & 
    \textcolor{gray}{$33.6$} & 
    \textcolor{gray}{$92.4$} & 
    \textcolor{gray}{$36.1$} & 
    \textcolor{gray}{$96.9$} & 
    \textcolor{gray}{$68.1$} & 
    \textcolor{gray}{$93.9$} & 
    \textcolor{gray}{$40.8$}\\
    \cline{1-19}
    WinCLIP~\cite{WinCLIP} & \ding{51} &  & $85.1$ & $64.6$ & $79.6$ & $56.8$ & $76.4$ & $48.9$ & $72.7$ & $27.3$ & $68.8$ & $24.2$ & $87.6$ & $65.7$ & $83.9$ & $57.8$ & $79.2$ & $49.3$ \\
    MuSc$^\dagger$~\cite{MuSc} & \ding{51} &  & $\textcolor{red}{\bf 97.2}$ & $\textcolor{red}{\bf 91.8}$ & $97.5$ & ${82.1}$ & $\textcolor{red}{\bf 96.9}$ & $\textcolor{red}{\bf 90.7}$ & $\textcolor{red}{\bf 97.9}$ & $77.5$ & $96.4$ & $82.2$ & $96.8$ & ${93.6}$ & $97.6$ & $93.9$ & ${97.2}$ & $87.4$ \\
    PatchEAD (CLIP) & \ding{51} & \ding{51} & $95.0$ & $88.9$ & $96.5$ & $79.1$ & $94.0$ & $82.7$ & $96.4$ & $71.2$ & $97.6$ & $86.4$ & $94.9$ & $88.0$ & $97.9$ & ${92.0}$ & $96.0$ & $84.0$ \\
    PatchEAD$^+$ (CLIP) & \ding{51} & \ding{51} & $94.4$ & $87.9$ & $96.3$ & $79.3$ & $94.3$ & $84.8$ & $96.5$ & $72.3$ & $97.7$ & $87.1$ & $95.3$ & $89.4$ & $97.7$ & $91.2$ & $96.0$ & $84.6$ \\
    PatchEAD (DINOv2) & \ding{51} & \ding{51} & $95.9$ & $91.3$ & $96.9$ & $87.5$ & $95.9$ & $89.3$ & ${97.4}$ & ${78.7}$ & $99.2$ & ${94.8}$ & $\textcolor{red}{\bf 97.2}$ & $\textcolor{red}{\bf 94.2}$ & $97.9$ & $92.6$ & $97.2$ & $89.8$ \\
    PatchEAD$^+$ (DINOv2) & \ding{51} & \ding{51} & $95.6$ & $91.4$ & $\textcolor{red}{\bf97.7}$ & $\textcolor{red}{\bf 91.2}$ & ${96.6}$ & ${89.6}$ & $97.0$ & $\textcolor{red}{\bf 80.0}$ & $\textcolor{red}{\bf 99.3}$ & $\textcolor{red}{\bf 97.7}$ & $96.4$ & $92.6$ & $\textcolor{red}{\bf98.3}$ & $\textcolor{red}{\bf 95.1}$ & $\textcolor{red}{\bf97.3}$ & $\textcolor{red}{\bf91.1}$ \\
\hline
\end{tabular}
}
}
\label{tab:zero_shot_image_comparison}
\end{table*}

\subsection{Cross-Patch Anomaly Scoring}
\label{ssec:Cross:Patch:Scoring}

We encode a query \(X_Q\) and \(N\) normal images \(\{X_P\}_{i=1}^{N}\) into
\(\mathbf{E}_Q\in\mathbb{R}^{L\times D}\) and \(\mathbf{E}_P\in\mathbb{R}^{N\times L\times D}\).
The definition of the normal image set varies according to the targeted task. In the few-shot setting, a fixed number $N$ of normal images is sampled. In contrast, the batch zero-shot setting considers a batch of $B$ samples, treating each sample as a query and using the remaining $B-1$ samples as normal references ({\em i.e.}, $N=B-1$). 

For a query patch \(j\), the cosine distance to the \(k\)-th patch of the \(i\)-th normal image is
\begin{equation}
\begin{aligned}
\mathbf{D}_{j,i,k}
&= 1 - \frac{\bigl\langle \mathbf{e}_Q^{\,j},\, \mathbf{e}_{P}^{\,i,k} \bigr\rangle}
{\lVert \mathbf{e}_Q^{\,j}\rVert_2 \, \lVert \mathbf{e}_{P}^{\,i,k}\rVert_2},
\end{aligned}
\end{equation}
A distance score for each normal image is represented by selecting the patch with the maximum distance in that image. Therefore, each query patch can obtain this distance score relative to each image written as follows:
\begin{equation}
\begin{aligned}
u_{j,i}
&= \max_{k\in\{1,\ldots,L\}} \mathbf{D}_{j,i,k},
\end{aligned}
\end{equation}

Next, we aim to aggregate $u_{j,i}$ for the $j^{th}$ patch as a patch-level anomaly score.
The selected images are the top $K_P$ highest-scoring images, denoted with the index set $\mathcal{I}_P$. 
The patch-level anomaly score for the $j^{th}$ patch is the average score across these $K$ images. 
\begin{equation}
\begin{aligned}
m_j
&= \frac{1}{K_P}\sum_{i\in \mathcal{I}_P} u_{j,i}.
\end{aligned}
\end{equation}
We set $K_P=1$ for the few-shot setting and $K_P=\max\bigl(1, 0.3N)$ for the batch zero-shot setting, {\em i.e.}, the top 30\% of images are used to compute patch-level anomaly scores. Each patch thus obtains the score based on the normal images that exhibit the most significant deviations.



Finally, the query image can derive an image-level score by averaging over its top $K_I$ patch scores:
\begin{equation}
\begin{aligned}
S = \frac{1}{K_I}\sum_{j\in \mathcal{I}_{I}} m_j,
\end{aligned}
\end{equation}
where $K_I=\max\bigl(1,\rho L\bigr)$, and $\mathcal{I}_I$ is the corresponding top-$K_I$ index set. 
This anomaly score $S$ determines the final prediction of the image.

\subsection{PatchEAD$^+$: Visual Prompt Enhancement} 
\label{ssec:robust}

We integrated optional alignment and masking features as part of the visual prompting techniques to enhance object alignment across images and reduce false positives from object edges and background noise. We refer to this {\em visual prompt-enhanced} version as PatchEAD$^+$ in the following experiments. Alignment reduces patch drift, and masking reduces background interference. Additional implementation details are provided in the supplementary material.

\noindent{\textbf{Alignment:}}
Object misalignment can cause patches to shift or rotate, resulting in higher anomaly scores during patch matching, especially at object edges. To address this, we use a feature-matching method from traditional image processing to align images through rotation and translation, ensuring consistent positioning and orientation between prompt and query images.

\noindent{\textbf{Masking:}}
We normalize the model’s final-layer attention scores to the [0, 1] range and apply a predefined threshold to binarize the result, producing a foreground mask that suppresses background regions. This approach mitigates the risk of misclassification caused by background noise.




\section{Experimental Evaluation}


{\bf Datasets:}
We conducted extensive evaluation experiments on {\bf seven} widely used industrial anomaly detection datasets to assess the performance of our PatchEAD framework: MVTec~\cite{mvtec}, VisA~\cite{visa}, MPDD~\cite{MPDD}, BTAD~\cite{BTAD}, KSDD~\cite{KSDD}, DAGM~\cite{DAGM}, and DTD-Synthetic~\cite{DTD-Synthetic}. These datasets collectively include over 50 different classes and defect types, providing a comprehensive testbed for real-world applications. PatchEAD was compared with {\bf five} state-of-the-art (SoTA) methods for few-shot anomaly detection: AprilGAN~\cite{AprilGAN}, AnomalyGPT~\cite{AnomalyCLIP}, PromptAD~\cite{PromptAD}, InCTRL~\cite{InCTRL}, and WinCLIP~\cite{WinCLIP}, as well as {\bf four} leading zero-shot/batch zero-shot methods: AnomalyCLIP~\cite{AnomalyCLIP}, AdaCLIP~\cite{AnomalyCLIP}, WinCLIP~\cite{WinCLIP}, and MuSc~\cite{MuSc}. These comprehensive comparisons demonstrate the training-free applicability of our framework across diverse datasets and detection settings.

\noindent{\bf Evaluation Metrics:}
For a comprehensive comparison with previous methods, we followed \cite{AnomalyCLIP} in using the Area Under the Receiver Operating Characteristic Curve (AUC) as an evaluation metric at both image and pixel levels. Additionally, we include image-level Average Precision (AP) and pixel-level Per-Region Overlap (PRO)~\cite{aupro} to better assess the model's performance on imbalanced datasets and its ability to localize anomalous regions.



\subsection{Implementation Details}



\noindent{\textbf{Model architecture:}}
For the reproduced CLIP-based model, we utilized the pre-trained ViT-B/16+ from OpenCLIP~\cite{openclip}, following the WinCLIP, MuSc, and AdaCLIP approaches to ensure a fair comparison. We also experimented with DINOv2~\cite{dinov2} and chose the ViT-B/14 distilled(with registers) variants as the backbone.

\noindent{\textbf{Few-shot setting:}}
We randomly sampled $N$ images from the training set of each category as prompts, where $N \in \{1, 2, 4\}$. To ensure robust results, we used three random seeds for sampling and reported the mean and standard deviation across trials.

\noindent{\textbf{Batch Zero-shot setting:}}
We evaluated batch zero-shot anomaly detection using batches of 64 unlabeled queries and processed most data in a single pass without replication.
This setup is consistent with the approach used in ACR~\cite{ACR} and MuSc~\cite{MuSc}.

\subsection{Details of Alignment and Masking}
\label{sec:alignment_masking}
We introduce both alignment and masking to the MVTec, VisA, KSDD, DAGM, and DTD-Synthetic datasets, and only masking is applied to MPDD and BTAD due to the objects being well-aligned (BTAD) or undergoing 3D rotation (MPDD).

{\noindent  \bf Alignment:} In the few-shot setting, we calculate the displacement and rotation between each prompt image and the query image. We then add the corrected images and their 180-degree rotations to the prompt list to ensure consistent positions and angles during patch matching. In the zero-shot setting, we use the first image as the query and apply the same process to all other images. For similarity calculation, we average the scores from the three images for each sample to determine the final anomaly score.

{\noindent  \bf Masking:} We extract the final attention layer and average all attention heads to get the [CLS] token’s attention scores for each patch. Given that the values vary substantially across model architectures and images, we first min–max normalize the attention scores to 0, 1. After normalizing, patches with scores above 0.05 are set to 1, and those below to 0.5. This reduces the incorrect activations in background areas.

\subsection{Few-/Batch Zero-Shot Detection Results}

Our experimental results cover both few-shot and batch zero-shot settings, aligning closely with prior works. Table~\ref{tab:few_shot_comparison} presents the \textbf{few-shot} results where the methods requiring training are shown in grey. While these training-based methods offer an upper bound on performance, they are often impractical for real-world use. Our proposed PatchEAD$^+$ (DINOv2) achieves remarkable performance in both image-level and pixel-level detection, approaching this upper bound.

PatchEAD (CLIP) achieves performance comparable to WinCLIP across different datasets and N-shot settings, demonstrating that patch-level information alone can deliver competitive results. Furthermore, using a more vision-centric backbone such as DINOv2, along with techniques like feature alignment and background removal to enhance the quality of visual prompts, further improves performance. These results demonstrate that, for anomaly detection tasks, both factors can provide benefits similar to those achieved with textual prompts.

\begin{figure}[t]
\centerline{
  \includegraphics[width=\linewidth]{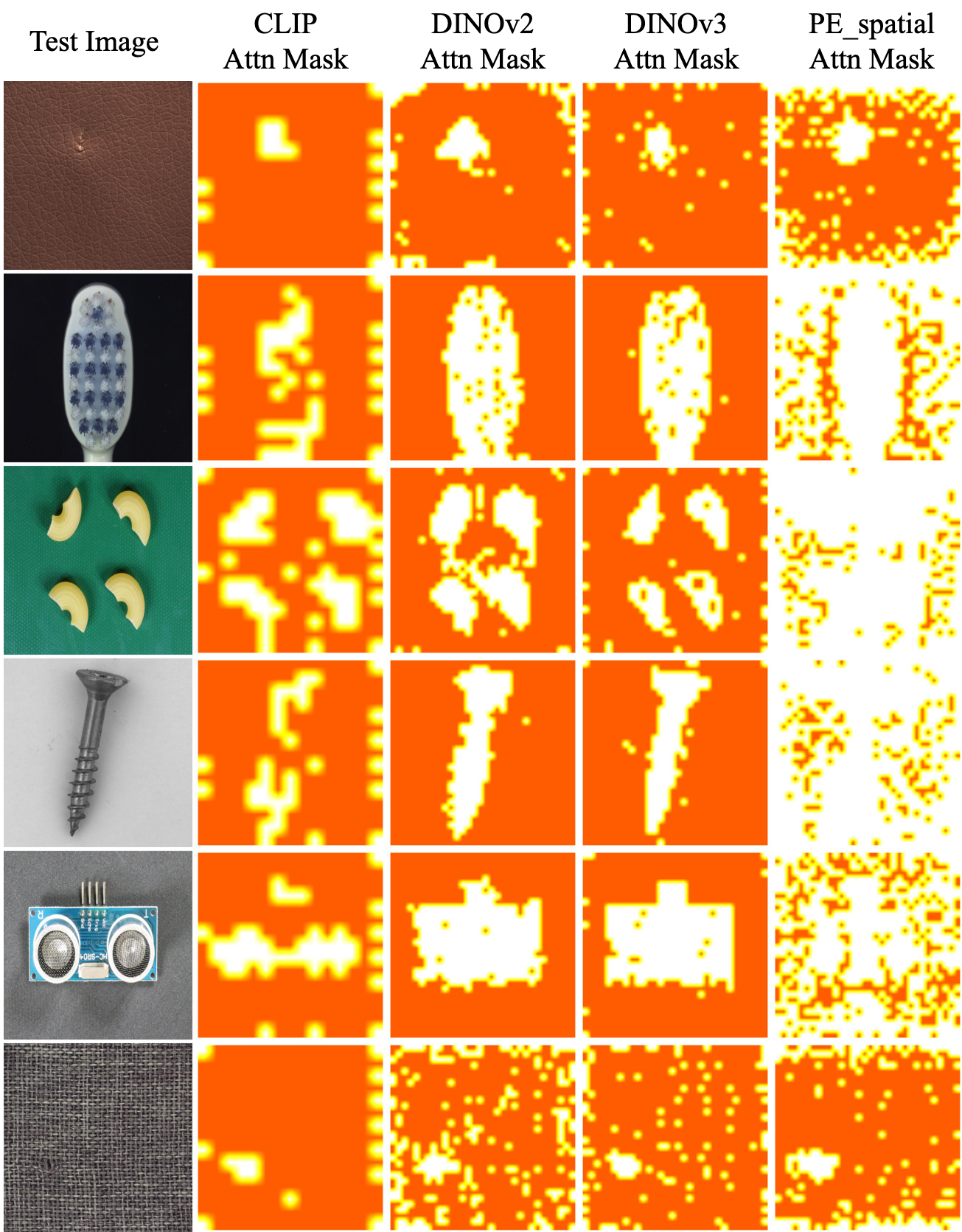} 
}
\caption{Visualization of attention masks from different backbones across normal and abnormal images, demonstrating the ability to down-weight noisy regions.}
\label{fig:attn_mask}
\end{figure}


\begin{table}[t]
\caption{
{\bf Ablation study on backbone selection,} evaluating alternative architectures: MAE~\cite{MAE}, BEiTv2~\cite{BEiTv2}, EVA-02~\cite{eva02}, PE\_spatial~\cite{perception_model}, and DINOv3~\cite{dinov3}, to assess their performance impact on image-level AUC (I-AUC) and pixel-level AUC (P-AUC) for MVTec and VisA datasets.
}
\label{tab:architecture_ablation}
\centerline{
\resizebox{\linewidth}{!}{
\setlength{\tabcolsep}{1mm}
\begin{tabular}{c|c|cc|cc}
\hline
\multirow{2}{*}{Setup} & \multirow{2}{*}{\makecell{PatchEAD\\backbone}} & \multicolumn{2}{c|}{MVTec~\cite{mvtec}} & \multicolumn{2}{c}{VisA~\cite{visa}} \\
\cline{3-6}
 &  & I-AUC & P-AUC & I-AUC & P-AUC \\
\hline
\multirow{7}{*}{Batch 0-shot} 
 & MAE~\cite{MAE} & $85.2$ & $94.3$ & $64.2$ & $80.6$ \\
 & BEiTv2~\cite{BEiTv2} & $84.9$ & $91.6$ & $74.2$ & $88.7$ \\
 & EVA-02~\cite{eva02} & $91.75$ & $91.3$ & $85.8$ & $92.9$ \\
 & DINOv2~\cite{dinov2} & $94.3$ & $95.9$ & $87.5$ & $96.9$ \\
 & PE\_spatial~\cite{perception_model} & $95.1$ & $96.7$ & $\mathbf{90.6}$ & $\mathbf{97.8}$ \\
 & DINOv3~\cite{dinov3} & $\mathbf{95.6}$ & $\mathbf{97.5}$ & $88.1$ & $97.0$ \\
\hline\hline
\multirow{7}{*}{1-shot} 
 & MAE~\cite{MAE} & $77.6_{\pm 0.4}$ & $90.7_{\pm 0.1}$ & $56.7_{\pm 2.5}$ & $76.5_{\pm 0.3}$ \\
 & BEiTv2~\cite{BEiTv2} & $86.8_{\pm 1.0}$ & $91.4_{\pm 0.4}$ & $75.9_{\pm 0.2}$ & $86.7_{\pm 0.1}$ \\
 & EVA-02~\cite{eva02} & $90.6_{\pm 1.6}$ & $91.5_{\pm 0.3}$ & $80.5_{\pm 0.6}$ & $91.6_{\pm 0.1}$ \\
 & CLIP~\cite{CLIP2021} & $91.6_{\pm 0.7}$ & $94.5_{\pm 0.3}$ & $81.6_{\pm 0.9}$ & $94.5_{\pm 0.2}$ \\
 & DINOv2~\cite{dinov2} & $\mathbf{93.6_{\pm 0.7}}$ & $96.0_{\pm 0.1}$ & $85.4_{\pm 0.3}$ & $\mathbf{97.3_{\pm 0.1}}$ \\
 & PE\_spatial~\cite{perception_model} & $93.1_{\pm 0.3}$ & $\mathbf{96.3_{\pm 0.2}}$ & $\mathbf{88.0_{\pm 0.7}}$ & $96.8_{\pm 0.1}$ \\
 & DINOv3~\cite{dinov3} & $91.3_{\pm 1.0}$ & $94.9_{\pm 0.2}$ & $81.2_{\pm 2.0}$ & $94.3_{\pm 0.1}$ \\
\hline
\multirow{7}{*}{2-shot} 
 & MAE~\cite{MAE} & $79.5_{\pm 0.5}$ & $91.8_{\pm 0.4}$ & $59.1_{\pm 0.7}$ & $78.2_{\pm 0.3}$ \\
 & BEiTv2~\cite{BEiTv2} & $87.7_{\pm 0.5}$ & $91.9_{\pm 0.2}$ & $77.3_{\pm 0.9}$ & $87.8_{\pm 0.3}$ \\
 & EVA-02~\cite{eva02} & $92.7_{\pm 0.6}$ & $91.8_{\pm 0.1}$ & $87.2_{\pm 0.4}$ & $92.5_{\pm 0.1}$ \\
 & CLIP~\cite{CLIP2021} & $91.6_{\pm 0.6}$ & $94.8_{\pm 0.1}$ & $85.3_{\pm 1.1}$ & $95.6_{\pm 0.1}$ \\
 & DINOv2~\cite{dinov2} & $\mathbf{95.0_{\pm 0.1}}$ & $96.3_{\pm 0.1}$ & $87.7_{\pm 0.3}$ & $\mathbf{97.7_{\pm 0.1}}$ \\
 & PE\_spatial~\cite{perception_model} & $93.1_{\pm 0.5}$ & $\mathbf{96.5_{\pm 0.2}}$ & $\mathbf{90.6_{\pm 0.4}}$ & $97.3_{\pm 0.1}$ \\
 & DINOv3~\cite{dinov3} & $93.5_{\pm 0.8}$ & $95.5_{\pm 0.1}$ & $85.4_{\pm 1.1}$ & $95.3_{\pm 0.1}$ \\
\hline
\multirow{7}{*}{4-shot} 
 & MAE~\cite{MAE} & $82.7_{\pm 0.4}$ & $93.5_{\pm 0.1}$ & $66.0_{\pm 1.1}$ & $80.4_{\pm 0.3}$ \\
 & BEiTv2~\cite{BEiTv2} & $89.1_{\pm 0.3}$ & $92.6_{\pm 0.1}$ & $78.5_{\pm 0.3}$ & $88.8_{\pm 0.2}$ \\
 & EVA-02~\cite{eva02} & $94.1_{\pm 1.1}$ & $92.5_{\pm 0.1}$ & $90.0_{\pm 0.5}$ & $93.0_{\pm 0.1}$ \\
 & CLIP~\cite{CLIP2021} & $94.1_{\pm 0.8}$ & $95.6_{\pm 0.1}$ & $86.7_{\pm 0.3}$ & $96.1_{\pm 0.1}$ \\
 & DINOv2~\cite{dinov2} & $\mathbf{96.0_{\pm 1.2}}$ & $96.8_{\pm 0.1}$ & $88.9_{\pm 0.1}$ & $\mathbf{98.0_{\pm 0.1}}$ \\
 & PE\_spatial~\cite{perception_model} & $95.1_{\pm 0.7}$ & $\mathbf{97.0_{\pm 0.1}}$ & $\mathbf{92.8_{\pm 0.1}}$ & $97.6_{\pm 0.1}$ \\
 & DINOv3~\cite{dinov3} & $94.4_{\pm 0.6}$ & $96.8_{\pm 0.1}$ & $89.4_{\pm 0.5}$ & $96.2_{\pm 0.2}$ \\
\hline
\end{tabular}
}}
\end{table}

\begin{table}[t]
\caption{
{\bf Ablation study} on the impact of image size using DINOv2, highlighting effects on patch receptive fields. I-AUC and P-AUC denote image- and pixel-level AUC, respectively. 0-shot indicates the batch zero-shot setting.
}
\label{tab:image_size_ablation}
\centerline{
\resizebox{\linewidth}{!}{
\setlength{\tabcolsep}{1mm}
\begin{tabular}{c|c|cc|cc}
\hline
\multirow{2}{*}{Setting} & \multirow{2}{*}{Image Size} & \multicolumn{2}{c|}{MVTec~\cite{mvtec}} & \multicolumn{2}{c}{VisA~\cite{visa}} \\
\cline{3-6}
 &  & I-AUC & P-AUC & I-AUC & P-AUC \\
\hline
0-shot & 224x224 & $90.7$ & $93.6$ & $82.8$ & $93.2$ \\
0-shot & 448x448 & $94.3$ & $95.9$ & $87.5$ & $96.9$ \\
0-shot & 672x672 & $\mathbf{94.7}$ & $\mathbf{96.4}$ & $\mathbf{89.1}$ & $\mathbf{97.5}$ \\
\hline
4-shot & 224x224 & $94.7_{\pm0.8}$ & $94.4_{\pm 0.1}$ & $85.1_{\pm 0.2}$ & $92.7_{\pm 0.1}$ \\
4-shot & 448x448 & $96.0_{\pm 1.2}$ & $\mathbf{96.8_{\pm 0.1}}$ & $88.9_{\pm 0.1}$ & $\mathbf{98.0_{\pm 0.1}}$ \\
4-shot & 672x672 & $\mathbf{96.2_{\pm 0.8}}$ & $96.7_{\pm 0.1}$ & $\mathbf{91.2_{\pm 0.4}}$ & $97.3_{\pm 0.1}$ \\
\hline
\end{tabular}
}
}
\end{table}

\begin{table}[t]
\caption{
{\bf Ablation study} on the impact of different modules in PatchEAD$^+$ (DINOv2), highlighting the contribution of each component. I-AUC and P-AUC denote image- and pixel-level AUC, respectively. 0-shot refers to the batch zero-shot setting.
}
\label{tab:module_ablation}
\centerline{
\resizebox{\linewidth}{!}{
\setlength{\tabcolsep}{1mm}
\begin{tabular}{c|cc|cc|cc}
\hline
\multirow{2}{*}{Setting} & \multicolumn{2}{c|}{Module} & \multicolumn{2}{c|}{MVTec~\cite{mvtec}} & \multicolumn{2}{c}{VisA~\cite{visa}} \\
\cline{2-7}
 & Alignment & Attn Mask & I-AUC & P-AUC & I-AUC & P-AUC \\
\hline
0-shot &  &  & $94.3$ & $95.9$ & $87.5$ & $96.9$ \\
0-shot & \checkmark &  & $94.8$ & $95.0$ & $85.7$ & $96.4$ \\
0-shot &  & \checkmark & $95.6$ & $\mathbf{96.2}$ & $91.8$ & $\mathbf{97.9}$ \\
0-shot & \checkmark & \checkmark & $\mathbf{95.9}$ & $95.6$ & $\mathbf{92.2}$ & $97.7$ \\
\hline
4-shot &  &  & $96.0_{\pm 1.2}$ & $96.8_{\pm 0.1}$ & $88.9_{\pm 0.1}$ & $\mathbf{98.0_{\pm 0.1}}$ \\
4-shot & \checkmark &  & $\mathbf{97.0_{\pm 0.8}}$ & $96.1_{\pm 0.1}$ & $90.1_{\pm 0.1}$ & $96.7_{\pm 0.1}$ \\
4-shot &  & \checkmark & $96.4_{\pm 0.8}$ & $\mathbf{96.9_{\pm 0.1}}$ & $91.8_{\pm 0.3}$ & $97.9_{\pm 0.1}$ \\
4-shot & \checkmark & \checkmark & $\mathbf{97.0_{\pm 0.8}}$ & $\mathbf{96.9_{\pm 0.1}}$ & $\mathbf{91.9_{\pm 0.3}}$ & $\mathbf{98.0_{\pm 0.1}}$ \\
\hline
\end{tabular}
}
}
\end{table}

The \textbf{batch zero-shot} results in Table~\ref{tab:zero_shot_image_comparison} show that our PatchEAD models outperform WinCLIP. PatchEAD$^+$ (DINOv2) achieves an average image-level AUC of 91.5\% and AP of 90.5\% across seven datasets, surpassing WinCLIP by 9.4\% in AUC and 9.9\% in AP. 
These results also outperform those of MuSc under the same batch zero-shot setting, even with MuSc’s advanced rescoring modules that leverage multi-level patch granularity. This demonstrates that higher-quality vision embeddings and enhanced vision prompts can significantly improve the cross-patch similarity framework.




\begin{figure}[t]
\centerline{
  \includegraphics[width=0.99\linewidth]{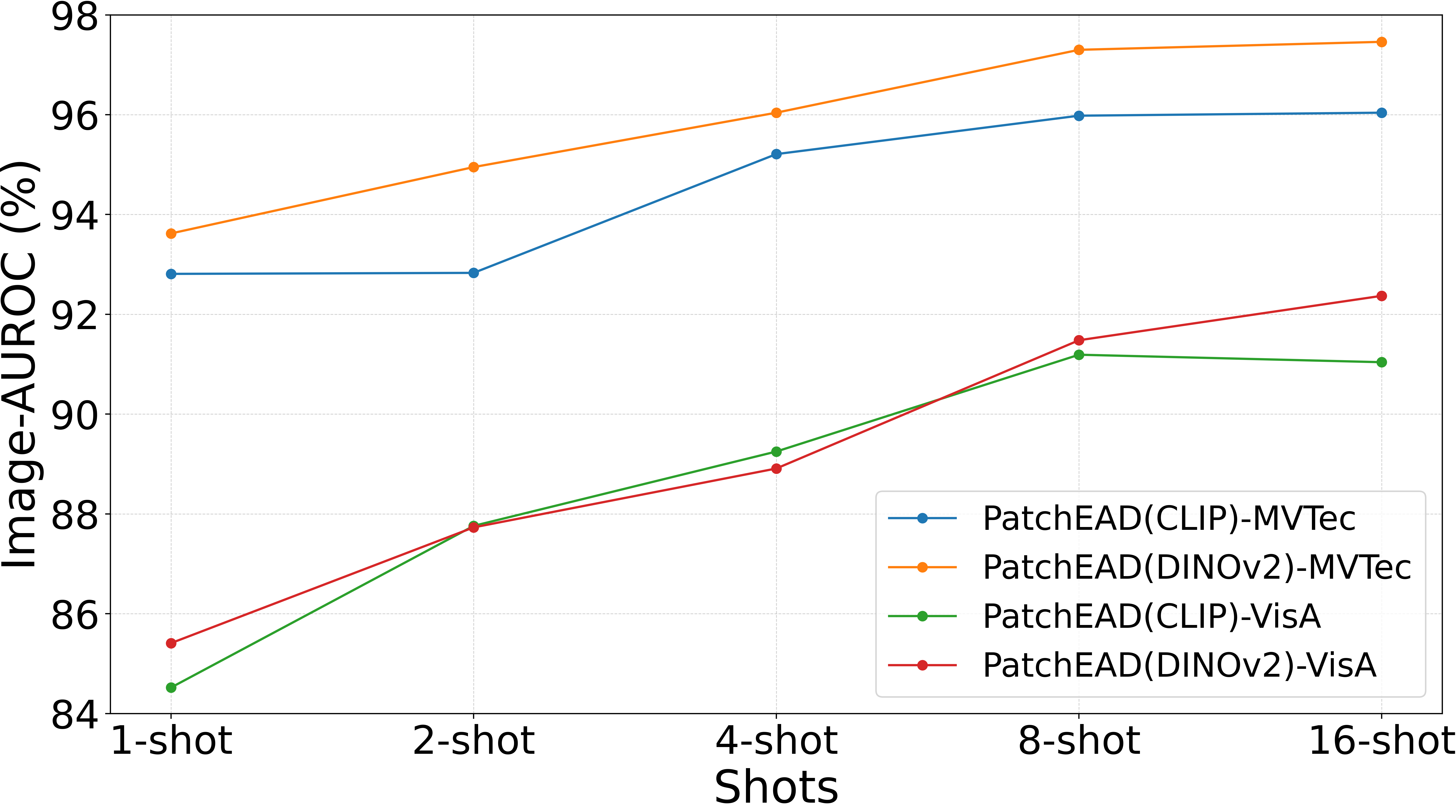} 
}
\caption{{\bf Ablation study} on the effect of the number of few-shot images in PatchEAD on image-level AUC for MVTec and VisA datasets.
}
\label{fig:shot_num_ablation}
\end{figure}


\subsection{Attention Mask Visualization}

We visualize the attention between the CLS token and patch tokens using four foundation models: CLIP, DINOv2, DINOv3, and PE\_spatial, as shown in Figure~\ref{fig:attn_mask}. Under the aforementioned masking rules, we observe that DINOv2 and DINOv3 exhibit the strongest capability in capturing foreground regions, while CLIP tends to focus on a few localized areas, and PE\_spatial produces comparatively smoother attention distributions. Nevertheless, all models consistently assign relatively high attention scores to defect regions. In our strategy, attention values below the threshold are down-weighted rather than set to zero, which improves the generalizability of the approach.

\subsection{Ablation Study}

{\textbf{Comparison of model architecture:}
We use MAE~\cite{MAE}, BEiTv2~\cite{BEiTv2}, EVA-02~\cite{eva02}, PE\_spatial~\cite{perception_model}, and DINOv3~\cite{dinov3} as our backbone models to analyze the impact of model size and architecture. As shown in Table~\ref{tab:architecture_ablation}, once the backbone reaches moderate capacity, this training-free, pure patch-similarity scheme produces consistently strong results in MVTec and VisA and different settings, highlighting the robust generalization of the framework.

\begin{figure*}[t]
\centerline{
  \includegraphics[width=\linewidth]{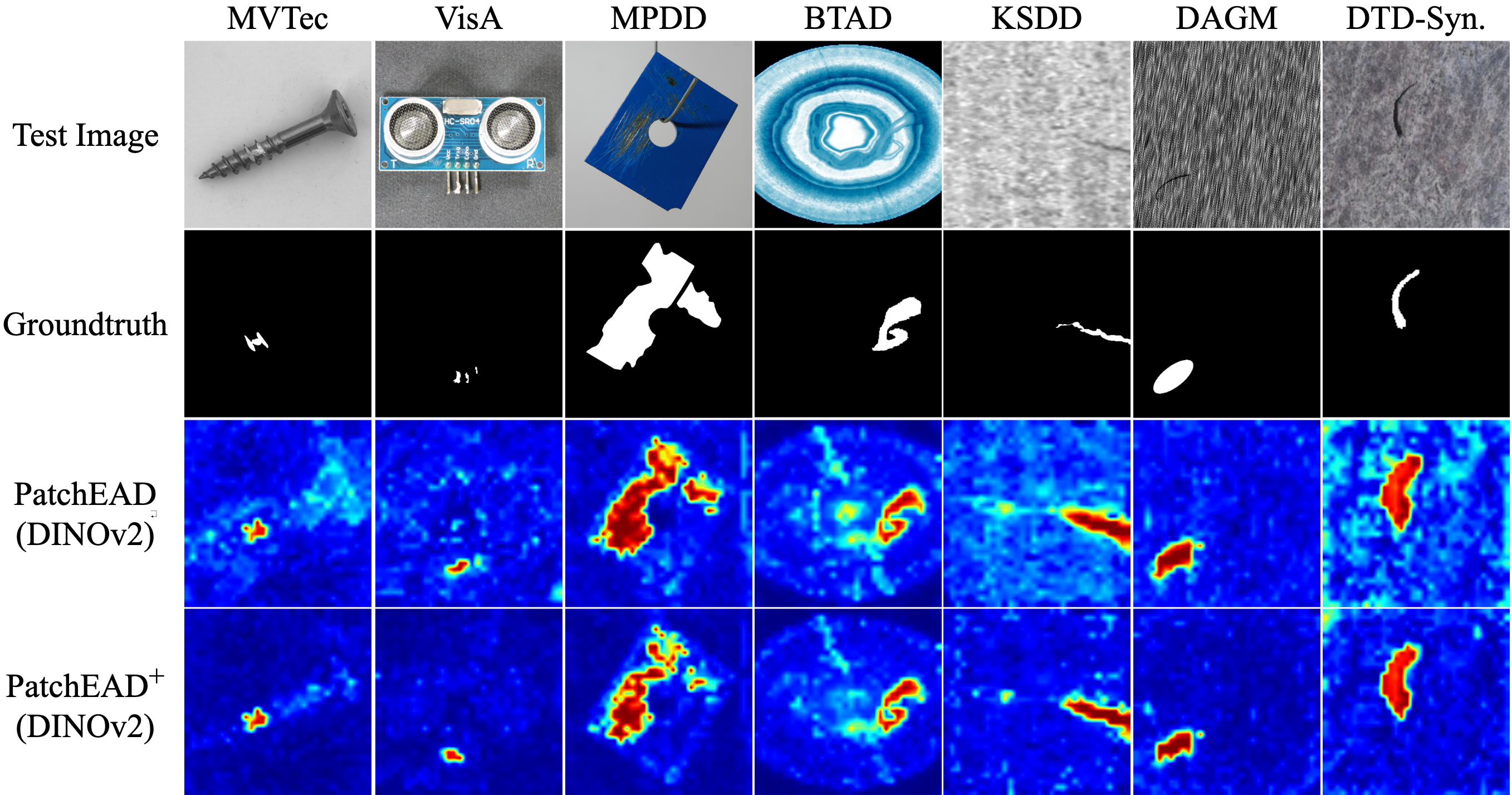} 
}
\caption{
{\bf Visualization of anomaly heatmaps generated by PatchEAD and PatchEAD$^+$} across seven industrial datasets, showing clearer and more precise defect outlines with reduced false positives near normal regions.
}
\label{fig:visualize_mask_align}
\vspace{-2mm}
\end{figure*}


\noindent{\textbf{Increasing number of few-shot normal images:}}
We investigated the impact of increasing the number of shots to determine whether patch information could be further enhanced with more reference images. As shown in Figure~\ref{fig:shot_num_ablation}, varying the number of inferencing shots from 1 to 16 led to a clear upward trend in image-level AUC scores. The most significant improvements were observed with 1-8 shots, with diminishing returns beyond that. Notably, increasing the number of shots beyond 16 yielded only marginal additional benefits. This experiment highlights a practical trade-off, emphasizing the need to balance few-shot data collection with adaptation efficiency.


\noindent{\textbf{Effects of increasing image size:}}
We investigated the impact of varying input image size on anomaly detection performance, which is a crucial factor influencing the model's receptive field, as our implementation maintains a constant patch size.
Experiments on the MVTec~\cite{mvtec} and VisA~\cite{visa} datasets are shown in Table~\ref{tab:image_size_ablation}. The results revealed consistently improved accuracies with $672 \times 672$ input size for both image-level and pixel-level AUC scores. 
Notably, using $448 \times 448$ already produced comparable accuracy to $672 \times 672$, offering a more efficient alternative. Larger image sizes correspond to smaller regions per patch, which may improve defect pattern recognition but increase inference time. Therefore, this parameter must balance computational efficiency with high accuracy for practical applications.



\noindent{\textbf{Effects of Alignment and Attention Mask:}}
In our ablation experiments on PatchEAD$^+$ using DINOv2, we evaluate the contributions of alignments and the attention mask as shown in Table~\ref{tab:module_ablation}. Results show that both components improve overall performance, and the best scores achieved when used together. Since we adopt a unified threshold setting rather than tuning attention mask thresholds for different object categories, pixel-level performance may vary and sometimes degrade depending on the object characteristics.





\subsection{Anomaly Mask Visualization}
Figure~\ref{fig:visualize_mask_align} presents visualizations of  PatchEAD, highlighting their ability to reduce false positives and enhance anomaly detection accuracy. PatchEAD$^+$ generates clearer and more precise defect outlines, minimizing false positives near normal regions. Further details of alignment and masking components in PatchEAD$^+$ are provided in the supplementary material.



\section{Conclusion}

Our work demonstrates the feasibility and effectiveness of a purely vision-based, patch-centric framework for industrial anomaly detection, in contrast to recent methods that rely on text prompts. Through extensive experiments in few-shot and batch zero-shot settings, PatchEAD consistently surpasses existing baselines, achieving strong performance without additional training or external normal libraries. Furthermore, PatchEAD$^+$ highlights critical design choices, such as targeted visual augmentations that stabilize the patch feature space and substantially enhance discriminative power. Importantly, our framework is backbone-agnostic, seamlessly integrating with any pre-trained vision encoder, making it immediately deployable and highly scalable across diverse industrial scenarios, and offering a practical solution for robust, real-world anomaly detection.




A limitation of our method is its reliance on visual patch information, which may miss non-visual anomalies, and small anomalous regions can produce indistinguishable patch features, reducing detection accuracy.

\textbf{Future works} includes refining the anomaly scoring function to better structure the feature space and enable more accurate extraction of patch-level and other nuanced information. Integrating multi-resolution techniques may further improve detection of anomalies across scales, enhancing both robustness and generalization in complex industrial settings.


{
    \small
    \bibliographystyle{ieeenat_fullname}
    \bibliography{main}
}
\appendix
\clearpage
\setcounter{page}{1}
\maketitlesupplementary
This supplementary material provides additional details to support the findings of our paper on training-free industrial anomaly detection with a patch-exclusive approach. $\S$~\ref{sec:dataset_details} includes information on the datasets used, detailing the types of multimodal data, preprocessing steps, and anomaly labeling. $\S$~\ref{sec:implementation_details} covers the architecture, feature extraction methods, and experimental setup, including hardware, software, and hyperparameters. $\S$~\ref{sec:preliminary} introduce a information theory to analyze the information relevant to the target task(anomaly detection) and the usage under mutli-modality scenario and show some experiments on text prompts compare to vision prompts. $\S$~\ref{sec:alignment_masking} demonstrate the detail of our alignment and masking strategy in PatchEAD$^+$. $\S$~\ref{sec:experiments_detail} present the evaluation process, including metrics and baseline comparisons. $\S$~\ref{sec:visualization} provides visualizations that demonstrate the effectiveness of our approach in detecting anomalies. Finally, $\S$~\ref{sec:case_study} presents a case study to showcase the practical application of our method in a real-world industrial scenario, highlighting its performance and potential challenges.


\section{Dataset Details}
\label{sec:dataset_details}

The experiments described in the main paper utilize seven widely recognized industrial anomaly detection datasets: MVTec~\cite{mvtec}, VisA~\cite{visa}, MPDD~\cite{MPDD}, BTAD~\cite{BTAD}, KSDD~\cite{KSDD}, DAGM~\cite{DAGM}, and DTD-Synthetic~\cite{DTD-Synthetic}. For datasets that provide a predefined training and testing split, we exclusively use the test data to evaluate the model’s performance. In cases where no such split is provided, we allocate 25\% of the normal images and all the anomalous images as the test set. For the KSDD dataset, each image is divided into three parts to better align its aspect ratio with the other datasets. Images containing a defect are labeled as anomalous, while those without a defect are considered normal. A detailed summary of the datasets, including key characteristics and splits, is presented in Table~\ref{tab:dataset_detail}.







\begin{table}[h]
\begin{center}
\resizebox{0.8\linewidth}{!}{
\setlength{\tabcolsep}{0.6mm}
    \footnotesize
    \begin{tabular}{cccccccc}
        \toprule
        \multirow{2}{*}{Dataset} & \multirow{2}{*}{Type}  & $\#$ of & \multicolumn{2}{c}{Samples} \\
        & & Categiries & Normal & Abnormal \\
        \midrule
        MVTec~\cite{mvtec}& Obj \& Texture &  15 & 467 & 1,258 \\
        VisA~\cite{visa} & Obj &  12 & 962 & 1,200 \\
        MPDD~\cite{MPDD} & Obj &  6 & 176 & 282 \\
        BTAD~\cite{BTAD} & Obj &  3 & 451 & 290 \\
        KSDD~\cite{KSDD} & Texture & 1 & 286 & 54 \\
        DAGM~\cite{DAGM} & Texture &  10 & 6,996 & 1,054 \\
        DTD-Synthetic~\cite{DTD-Synthetic} & Texture &  12 & 357 & 947 \\
        \bottomrule
    \end{tabular}
}
\end{center}
    \caption{Anomaly Detection Datasets in Industrial Domain.}
    \label{tab:dataset_detail}
\end{table}

\section{Implementation Details}
\label{sec:implementation_details}


Across all experiments, we adopt base-size backbones to keep parameter counts comparable. Our CLIP variant uses \texttt{ViT-B-16-plus-240} pretrained on \texttt{laion400m\_e32} from OpenCLIP~\cite{openclip}, with inputs resized to $240\times240$ to match the checkpoint. The DINOv2 variant employs \texttt{dinov2\_vitb14\_reg} from the official codebase~\cite{dinov2} at its default $448\times448$ resolution. For PE\_spatial, we load \texttt{PE-Spatial-B16-512}; for DINOv3, we use the distilled \texttt{ViT-B/16}. These settings keep each method close to its standard configuration and enable a fair comparison across architectures.

\section{Multimodal Information Analysis}
\label{sec:preliminary}

\begin{figure}[t]
\centering
\begin{subfigure}[t]{0.61\linewidth}
  \centering
  \includegraphics[width=\linewidth]{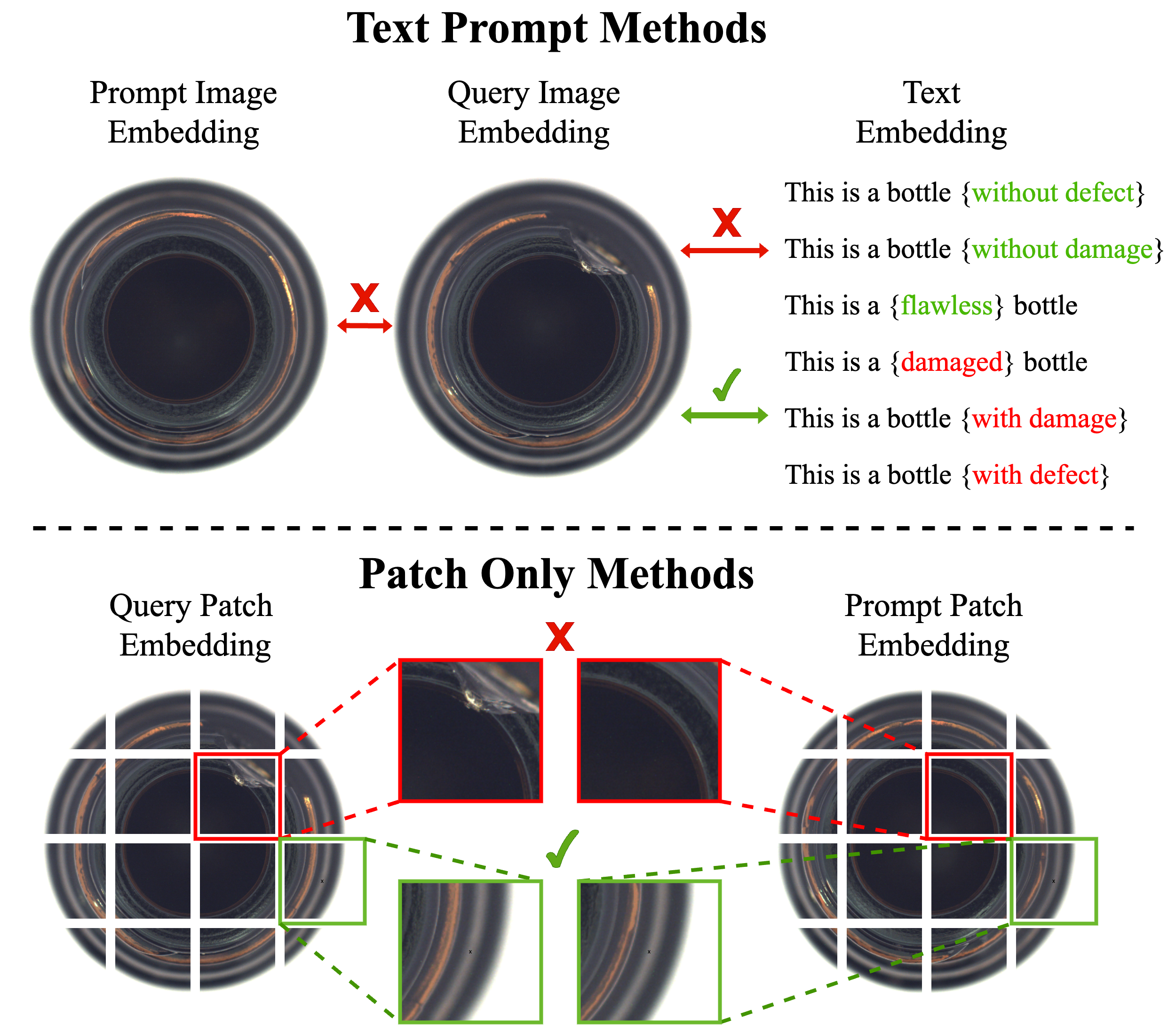}
  \caption{Overview of Patch vs. Text}\label{fig:teaser:a}
\end{subfigure}
\hfill
\begin{subfigure}[t]{0.37\linewidth}
  \centering
  \includegraphics[width=\linewidth]{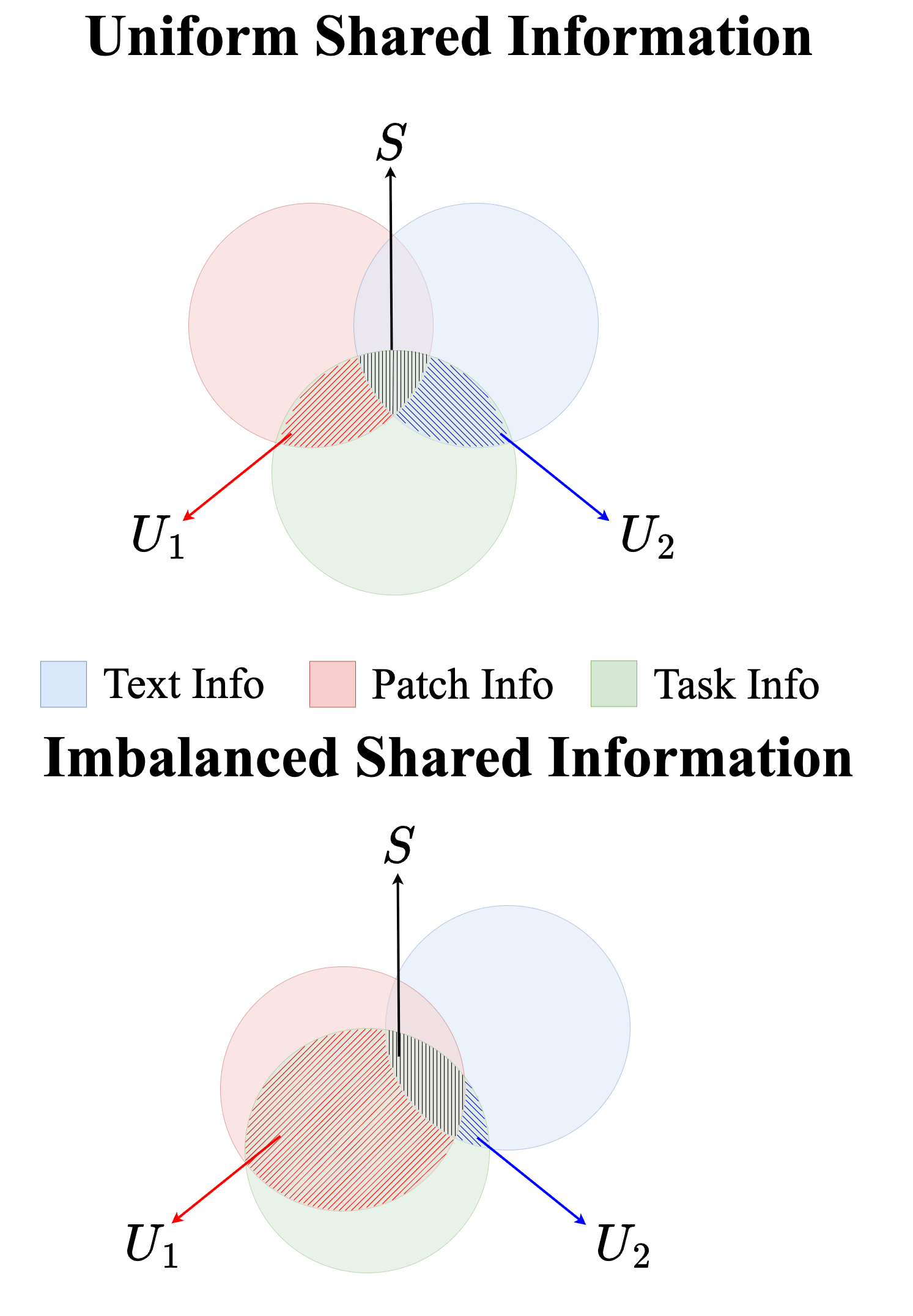}
  \caption{Assumption}\label{fig:teaser:b}
\end{subfigure}

\caption{(a) Existing anomaly detection methods using image–text similarity often assume clean separation via simple positive/negative prompts. Our \textbf{patch-only} method detects anomalies by comparing query patches with normal prompt patches, leveraging richer context. (b) Considering shared information $S$ and modality-unique information $U_1, U_2$, task relevance varies; we argue patch-level information outweighs text information for industrial defect detection.}
\label{fig:teaser}
\end{figure}

Following the formulation in \cite{liang2023factorized} for handling multimodal information, we consider two input modalities, 
$X_i (i \in \{1, 2\})$, annotated with a label variable $Y$.
The joint distribution of these inputs and $Y$ can be expressed using conditional mutual information:
\begin{equation}
I(X_1, X_2, Y) = S + U_1 + U_2, 
\end{equation}
where $S = I(X_1;X_2;Y)$ denotes the task-relevant shared information; $U_1 = I(X_1;Y|X_2)$ and $U_2 = I(X_2;Y|X_1)$ denote the unique information of modalities $X_1$ and $X_2$, respectively. Figure~\ref{fig:teaser} provides a concept plot of the imbalanced shared {\em v.s.} unique information provided in different modalities. Our hypothesis is that patch-level information should be more important over the text information for industrial defect detection. Our aim is to uncover the imbalanced distribution of $U_1$ and $U_2$ across different modalities of texts, patches, and whole images.



Loss functions in large multi-modality models, such as CLIP~\cite{CLIP2021}, are often optimized to capture shared information across modalities. Consequently, other anomaly detection methods tend to add additional modules to enhance information. However, these additional modules may introduce overlapping information between modalities. Our focus is to examine the unique information provided by each modality ($U_1$ and $U_2$). 

The unique information ($U_1$ and $U_2$) can be challenging to compute directly, so we approximate it using an upper-bound, $I_{CLUB}$, as proposed in contrastive learning~\cite{cheng2020club}. When $I_{CLUB}$ is small, the unique information in a modality is expected to be limited. For instance, we can estimate the unique information $U_1$ as follows: 

\begin{equation}
\scalebox{0.8}{$
    U_1 = I(X_1;Y|X_2) \leq I_{CLUB}(X_1;Y|X_2),
$}
\end{equation} 

where $I_{CLUB}$ is given by:

\begin{equation}
\scalebox{0.8}{$
    I_{CLUB}(X_1;Y|X_2)= E[f^*(X_1, Y^+|X_2)] - E[f^*(X_1, Y^-|X_2)], \\
\label{eq:I_CLUB}
$}
\end{equation}
with $f^*$ as the optimal critic from a contrastive pre-trained network. $I_{CLUB}$ represents the expected difference between positive and negative samples, $X^+$ and $X^-$ for a given $X$.
By applying an optimal augmentation assumption~\cite{tsai2021self}, we replace $Y$ with an augmented variable $X'$ of $X$, assuming that only task-relevant information is shared between $X$ and $X'$. $I_{CLUB}$ can then be calculated via this adjustment as:

\begin{equation}
\scalebox{0.8}{$
    I_{CLUB}(X_1;Y|X_2) = E[f^*(X_1, X_1'^+|X_2)] - E[f^*(X_1, X'^-|X_2)]. \\
\label{eq:I_CLUB_Final}
$}
\end{equation} 
If the expected anomaly scores between positive and negative samples are not significantly different, this upper bound will be small, suggesting limited unique information. Minimal differences in anomaly scores may also result if $X_2$ provides sufficient information, thus reducing the reliance on $X_1'^+$ and $X_1'^-$.


Considering $X_1$ for the patch modality and $X_2$ as the text modality, we use Eq.~\eqref{eq:I_CLUB_Final} to derive the upper bounds for $U_1$ and $U_2$. The upper bound for $U_2$, $I_{CLUB}(X_2;Y|X_1)$, is expected to be small because augmenting text prompts results in minimal changes in anomaly scores once patches are already included. On the other hand, the upper bound for $U_1$, $I_{CLUB}(X_1;Y|X_2)$, can be large, as augmentations like rotation, cropping, or noise perturbations cause noticeable variations in the final anomaly scores. This larger upper bound for $U_1$ arises because the text prompts often fail to capture details about missing objects or defects, which are better represented in the patch modality.

\begin{figure*}[t]
\centering
\begin{subfigure}[t]{0.24\textwidth}
  \centering
  \includegraphics[width=\linewidth]{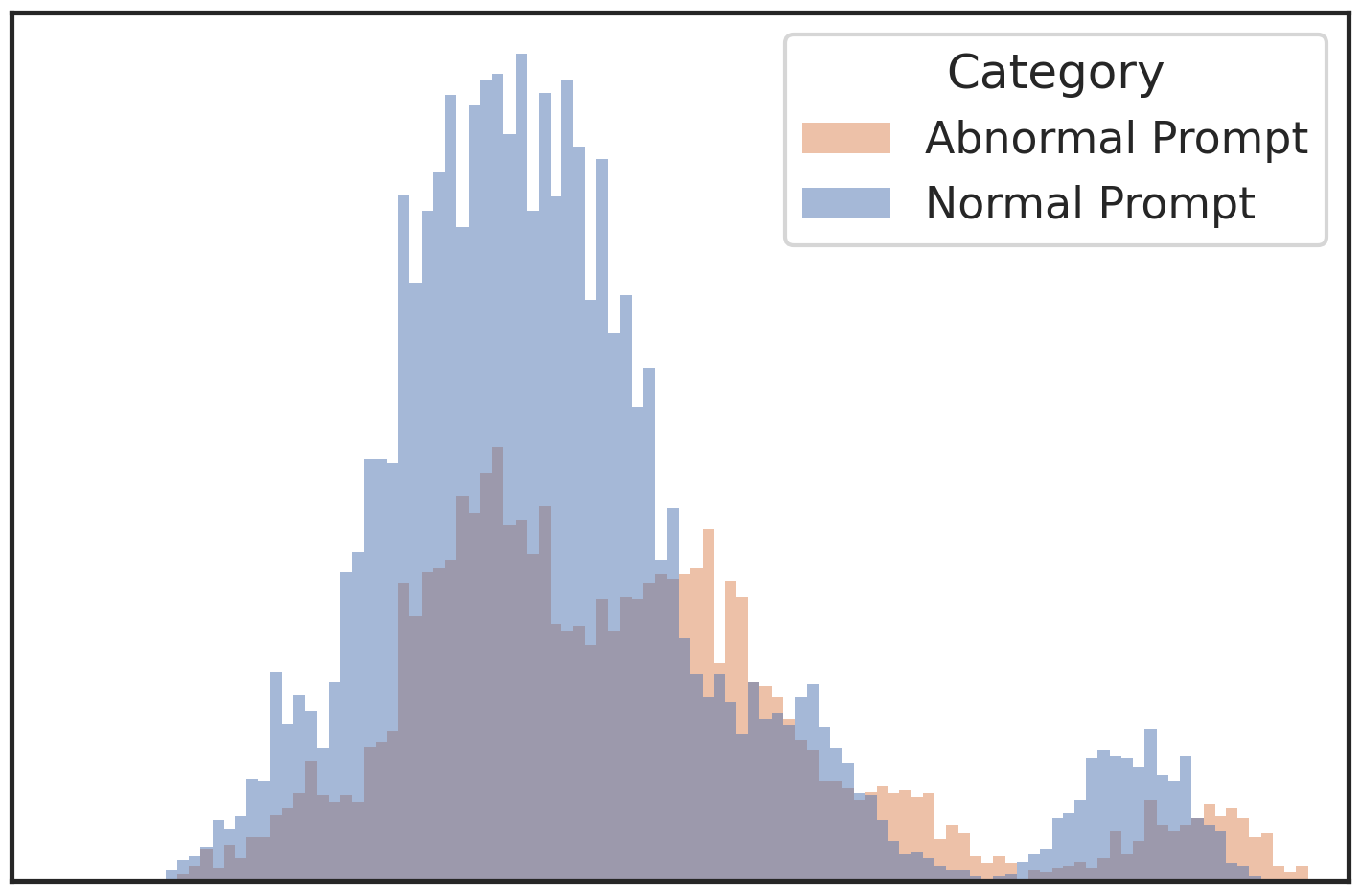}
  \caption{Normal images}\label{fig:text_patch_comparison:a}
\end{subfigure}
\hfill
\begin{subfigure}[t]{0.24\textwidth}
  \centering
  \includegraphics[width=\linewidth]{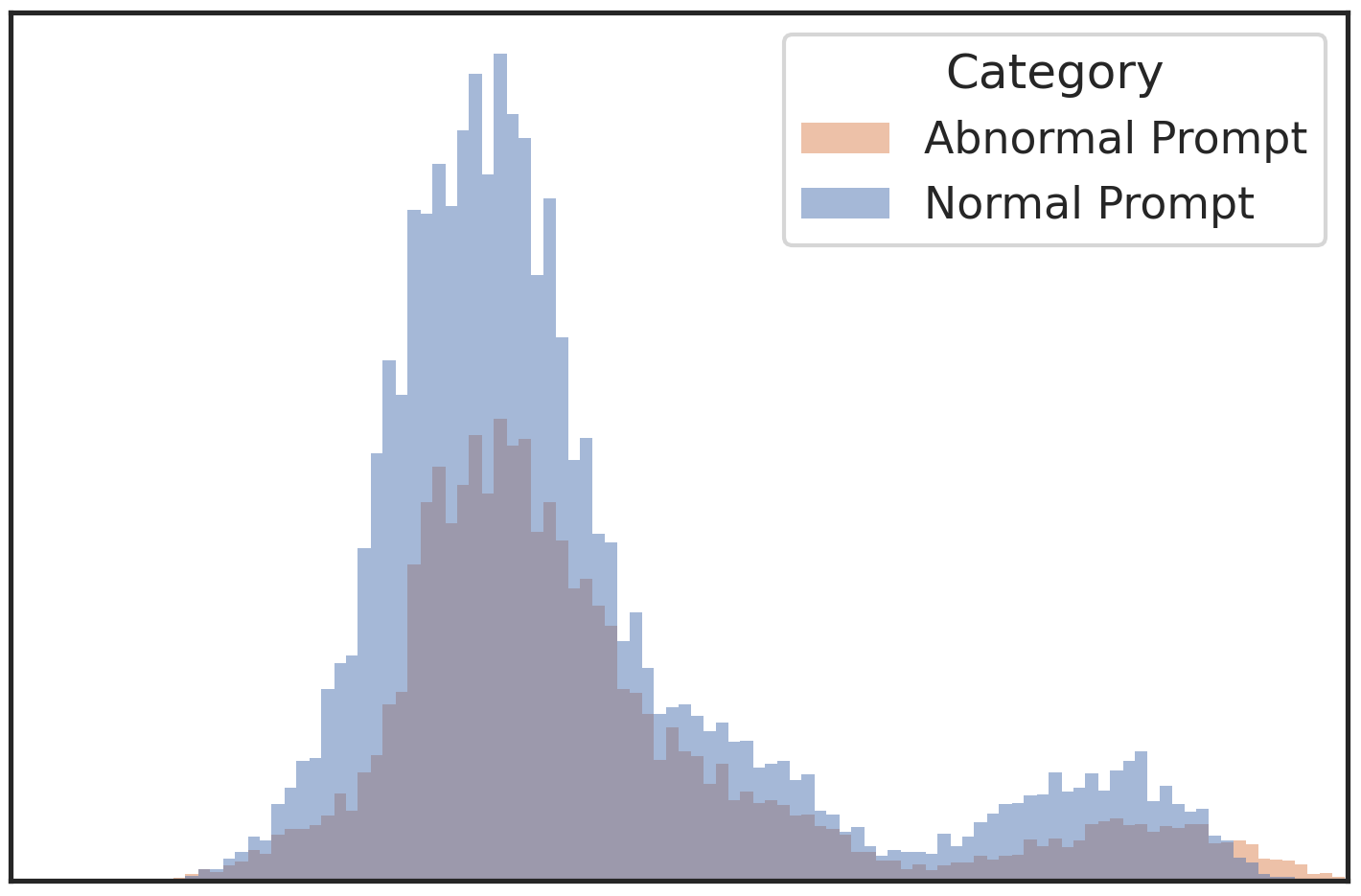}
  \caption{Abnormal images}\label{fig:text_patch_comparison:b}
\end{subfigure}
\begin{subfigure}[t]{0.24\textwidth}
  \centering
  \includegraphics[width=\linewidth]{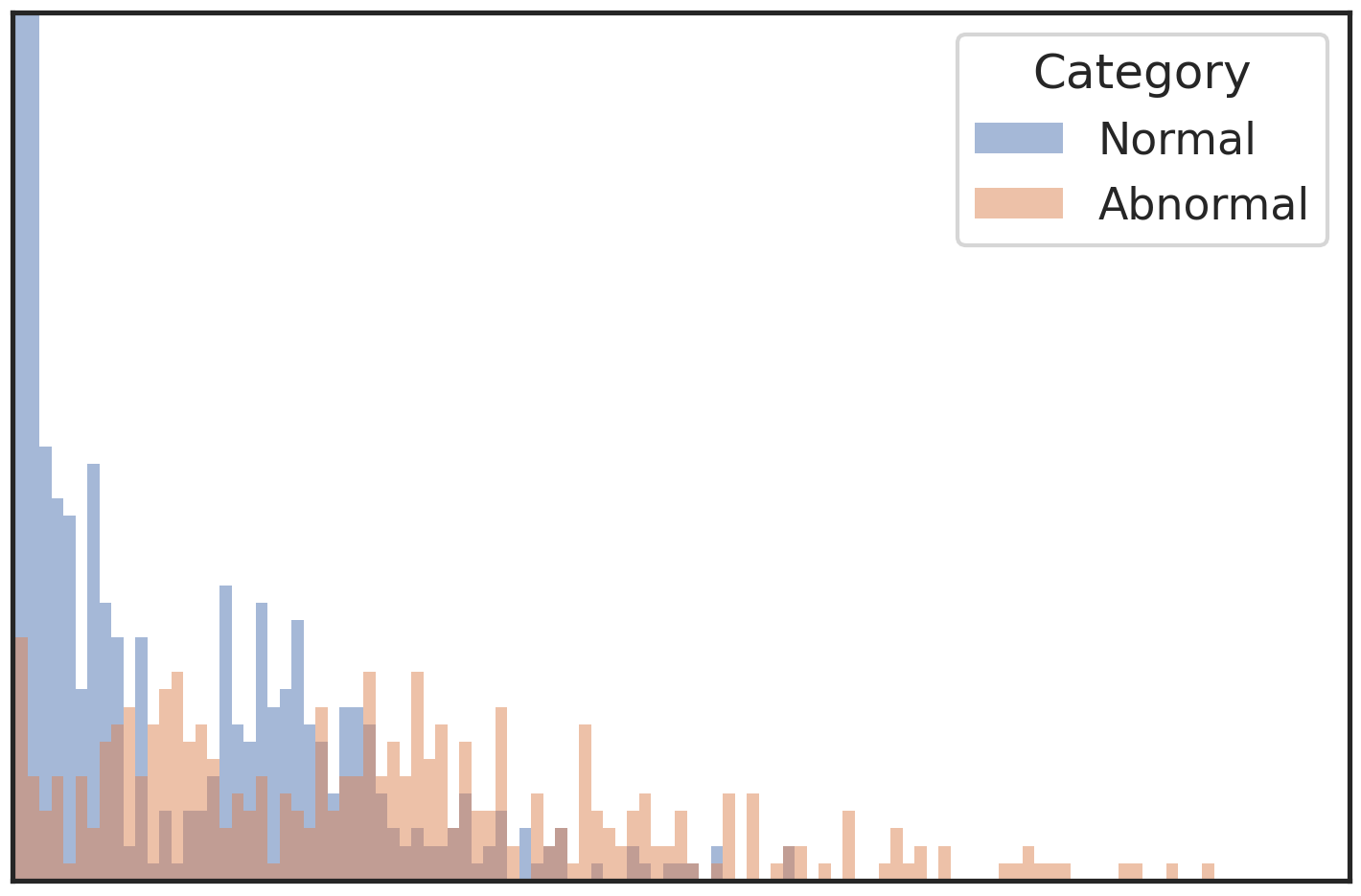}
  \caption{Text prompts}\label{fig:text_patch_comparison:c}
\end{subfigure}
\hfill
\begin{subfigure}[t]{0.24\textwidth}
  \centering
  \includegraphics[width=\linewidth]{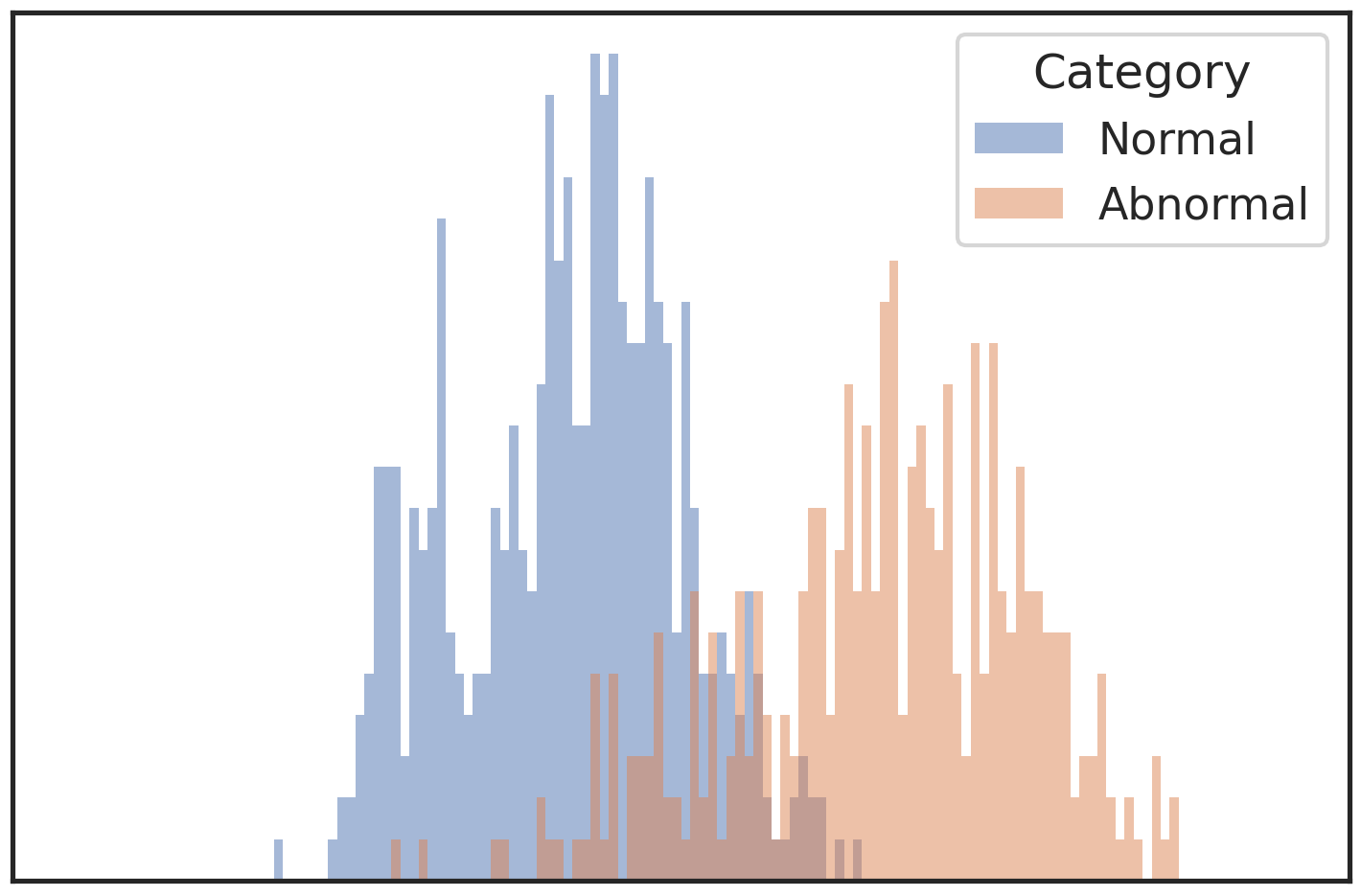}
  \caption{Patch prompts}\label{fig:text_patch_comparison:d}
\end{subfigure}

\caption{\textbf{Histogram plots} summarizing experimental results on the MVTec dataset.
\textbf{(a)} and \textbf{(b)} depict the \textbf{image–text similarity scores} for normal and abnormal images under various text prompts. Ideally, normal images should exhibit high similarity with their corresponding normal prompts and lower similarity when paired with abnormal prompts. \textbf{(c)} and \textbf{(d)} show the \textbf{anomaly scores} computed using text and patch prompts, respectively.}
\label{fig:text_patch_comparison}
\end{figure*}


\subsection{Comparison of Text {\em vs.} Patch Information}

Figure~\ref{fig:text_patch_comparison}(a,b) compares the {\bf normal {\em vs.} abnormal text prompts} on the MVTec dataset.  These plots explore how different prompts affect the image-text similarity score histograms for normal and abnormal samples. The information upper bound of each modality, as described in Eq.~\eqref{eq:I_CLUB_Final}, is estimated based on the expected score difference between positive and negative augmentations. By comparing the distributions of scores for positive and negative prompts in panels (a) and (b), we illustrate how the text information upper bound impacts the anomaly detection performance. 

Figure~\ref{fig:text_patch_comparison}(c,d) compares the {\bf text {\em vs.} patch modalities} by using the histogram of anomaly scores on the MVTec dataset. (c) plots the anomaly score histogram using WinCLIP, a model featuring vision-text multimodal capability. The scores were derived from the similarity between the image and text prompt tokens. 
(d) plots the histogram using our PatchEAD, where only patch information is used. 
Observe that (d) exhibits a more separable space than (c), suggesting more discriminative results with only patch information.

Results show that the positive and negative prompts yield a highly overlapped distribution regardless of normal or abnormal images. The upper bound of text information is limited, leading to inferior discriminative power than patches. 


\section{Details of Alignment and Masking}
\label{sec:alignment_masking}
We introduce both alignment and masking to MVTec, VisA, KSDD, DAGM, and DTD-Synthetic datasets, and only masking applied on MPDD, and BTAD due to the object is well-aligned(BTAD) or 3D rotation(MPDD).

{\noindent  \bf Alignment:} In the few-shot setting, we calculate the displacement and rotation between each prompt image and the query image. We then add the corrected images and their 180-degree rotations to the prompt list to ensure consistent positions and angles during patch matching. In the zero-shot setting, we use the first image as the query and apply the same process to all other images. For similarity calculation, we average the scores from the three images for each sample to determine the final anomaly score.

{\noindent  \bf Masking:} We extract the final attention layer and average all attention heads to get the [CLS] token’s attention scores for each patch. After normalizing, patches with scores above 0.05 are set to 1, and those below to 0.5. This reduces the incorrect activations in background areas.

\section{Experimental Details}
\label{sec:experiments_detail}

Our experiments entail few-shot and zero-shot settings with various datasets and thus only present the average results in the main manuscript under the page limitation. We report the few-shot results with each category in the MVTec and ViSA datasets in Table~\ref{tab:mvtec_few_shot} and Table~\ref{tab:visa_few_shot}. The zero-shot results are reported in Table~\ref{tab:mvtec_visa_mpdd_zero_shot} using seven datasets.

When we reveal the category-wise AUC in the few-shot setting, we observe the capsule, screw, and transistor in the MVTec dataset are harder categories with lower image-level and pixel-level AUC. Using DINOv2 significantly improves PatchEAD in these categories compared to CLIP, highlighting the benefits of vision purpose pretraining for the patch-only method, even when performed in an unsupervised manner. This improvement can consistently be observed in 1-, 2-, and 4-shot results, especially on image-level AUC. PatchEAD$^+$ strengthening alignment and masking mainly facilitates the screw category in different few-shot settings. 

Similarly, the few-shot image-level AUC attains relatively significant improvements on specific categories such as capsules, fryum, macaroni1, and pcb2 when comparing DINOv2 and CLIP in the VisA dataset. The improvement using PatchEAD$^+$ is eminent in the fryum, macaroni1, and pcb2 categories on both image-level and pixel-level AUC over different few-shot settings. 
The enhanced categories are generally aligned between few-shot and zero-shot settings. Our proposed PatchEAD using only patch information complements these hard cases with delicate visual appearances.

\section{Visualization Results}
\label{sec:visualization}

Additionally, to demonstrate the defect localization capability of our proposed PatchEAD framework, we conducted visualization on all test datasets. Specifically, we linearly interpolated the predicted patch anomaly scores to match the size of the original images, followed by normalization, resulting in a heatmap for each image. In the few-shot setting, we used the MVTec and VisA datasets for visualization, with results shown in Figures~\ref{fig:mvtec_few} and ~\ref{fig:visa_few}. In the zero-shot setting, we conducted experiments using seven datasets, with results presented in Figures ~\ref{fig:mvtec_zero},~\ref{fig:visa_zero},~\ref{fig:mpdd_zero},~\ref{fig:btad_zero},~\ref{fig:ksdd_zero},~\ref{fig:dagm_zero},~\ref{fig:dtd_synthetic_zero}. These test datasets encompass a wide variety of object types, and the visualization results indicate that our method, even without training, can accurately localize various defects under different settings, demonstrating the framework's strong flexibility and generalization capability.

\section{Case Study}
\label{sec:case_study}

We conducted a visual comparison to evaluate the effectiveness of alignment and masking in our method. As shown in Figure~\ref{fig:case_study}, masking reduces the weight of background regions, thereby decreasing the likelihood of misclassification. Meanwhile, alignment addresses variations in object orientation and displacement by aligning objects, enabling the comparison of similar contours and significantly reducing misclassification at object edges.

\begin{table*}[ht]
    \begin{center}
    \resizebox{0.8\linewidth}{!}{
    \setlength{\tabcolsep}{1mm}
    \begin{tabular}{ccccc|cccc}
        \toprule
        \multirow{3}{*}{Category} & \multicolumn{4}{c}{\textbf{Image-Level AUC}} & \multicolumn{4}{c}{\textbf{Pixel-Level AUC}}\\
        \cmidrule(lr){2-5} \cmidrule(lr){6-9}
          & PatchEAD & PatchEAD & PatchEAD+ & PatchEAD+ & PatchEAD & PatchEAD & PatchEAD+ & PatchEAD+ \\
        \cmidrule(lr){2-5} \cmidrule(lr){6-9}
          & ViT-B/16+ & DINOv2-B & ViT-B/16+ & DINOv2-B & ViT-B/16+ & DINOv2-B & ViT-B/16+ & DINOv2-B \\
        \midrule
        \multicolumn{9}{c}{\textbf{1-shot}} \\
        \midrule
        bottle      & $98.2_{\pm 0.7}$ & $98.8_{\pm 1.6}$ & $98.5_{\pm 0.7}$ & $99.7_{\pm 0.4}$ & $97.9_{\pm 0.1}$ & $98.3_{\pm 0.1}$ & $98.0_{\pm 0.1}$ & $98.1_{\pm 0.1}$\\
        cable       & $90.4_{\pm 2.0}$ & $90.8_{\pm 1.9}$ & $88.8_{\pm 2.0}$ & $92.5_{\pm 1.1}$ & $93.9_{\pm 0.4}$ & $93.2_{\pm 0.5}$ & $93.9_{\pm 0.3}$ & $93.4_{\pm 0.4}$\\
        capsule     & $80.8_{\pm 12.0}$ & $82.4_{\pm 13.3}$ & $78.2_{\pm 11.8}$ & $85.9_{\pm 11.2}$ & $90.6_{\pm 0.7}$ & $97.4_{\pm 0.2}$ & $93.4_{\pm 0.6}$ & $98.0_{\pm 0.2}$\\
        carpet      & $99.9_{\pm 0.1}$ & $98.5_{\pm 1.4}$ & $100.0_{\pm 0.0}$ & $99.6_{\pm 0.3}$ & $99.3_{\pm 0.1}$ & $98.7_{\pm 0.1}$ & $99.3_{\pm 0.0}$ & $98.7_{\pm 0.1}$\\
        grid        & $99.4_{\pm 0.5}$ & $99.1_{\pm 0.6}$ & $99.7_{\pm 0.2}$ & $97.7_{\pm 0.6}$ & $98.2_{\pm 0.1}$ & $98.7_{\pm 0.1}$ & $98.2_{\pm 0.1}$ & $98.7_{\pm 0.2}$\\
        hazelnut    & $99.4_{\pm 0.2}$ & $98.6_{\pm 0.8}$ & $99.7_{\pm 0.2}$ & $99.4_{\pm 0.3}$ & $98.8_{\pm 0.2}$ & $98.9_{\pm 0.1}$ & $98.2_{\pm 0.1}$ & $99.4_{\pm 0.0}$\\
        leather     & $100.0_{\pm 0.0}$ & $99.8_{\pm 0.2}$ & $100.0_{\pm 0.0}$ & $100.0_{\pm 0.0}$ & $99.3_{\pm 0.0}$ & $98.8_{\pm 0.1}$ & $99.4_{\pm 0.0}$ & $99.0_{\pm 0.1}$\\
        metal\_nut  & $99.9_{\pm 0.1}$ & $98.8_{\pm 0.9}$ & $98.3_{\pm 0.3}$ & $99.7_{\pm 0.2}$ & $87.2_{\pm 0.5}$ & $95.8_{\pm 0.4}$ & $90.5_{\pm 0.3}$ & $95.4_{\pm 0.5}$\\
        pill        & $96.3_{\pm 0.2}$ & $91.4_{\pm 5.1}$ & $93.8_{\pm 0.6}$ & $93.7_{\pm 3.6}$ & $93.8_{\pm 0.4}$ & $93.8_{\pm 0.6}$ & $93.0_{\pm 0.5}$ & $94.1_{\pm 0.7}$\\
        screw       & $62.8_{\pm 5.5}$ & $77.4_{\pm 4.2}$ & $78.5_{\pm 1.3}$ & $80.6_{\pm 3.7}$ & $92.5_{\pm 0.9}$ & $97.2_{\pm 0.2}$ & $94.8_{\pm 0.6}$ & $97.9_{\pm 0.1}$\\
        tile        & $99.5_{\pm 0.3}$ & $99.8_{\pm 0.1}$ & $98.4_{\pm 0.7}$ & $100.0_{\pm 0.0}$ & $95.4_{\pm 0.2}$ & $94.4_{\pm 0.2}$ & $95.4_{\pm 0.1}$ & $95.0_{\pm 0.2}$\\
        toothbrush  & $95.2_{\pm 1.0}$ & $94.0_{\pm 4.4}$ & $94.9_{\pm 0.9}$ & $94.4_{\pm 3.1}$ & $97.7_{\pm 0.3}$ & $98.8_{\pm 0.2}$ & $98.1_{\pm 0.2}$ & $98.2_{\pm 0.3}$\\
        transistor  & $75.6_{\pm 7.5}$ & $76.3_{\pm 7.0}$ & $81.5_{\pm 4.6}$ & $82.0_{\pm 6.1}$ & $86.7_{\pm 1.7}$ & $86.7_{\pm 1.4}$ & $84.5_{\pm 1.7}$ & $85.8_{\pm 1.5}$\\
        wood        & $98.8_{\pm 0.4}$ & $99.3_{\pm 0.2}$ & $99.1_{\pm 0.4}$ & $100.0_{\pm 0.0}$ & $95.2_{\pm 0.2}$ & $92.1_{\pm 0.9}$ & $94.9_{\pm 0.2}$ & $93.4_{\pm 1.0}$\\
        zipper      & $94.1_{\pm 1.0}$ & $99.2_{\pm 0.2}$ & $92.2_{\pm 1.5}$ & $99.7_{\pm 0.1}$ & $90.7_{\pm 0.8}$ & $96.9_{\pm 0.4}$ & $91.9_{\pm 0.4}$ & $97.1_{\pm 0.3}$\\
        \midrule
        \textbf{Average} & $92.8_{\pm 0.9}$ & \underline{$93.6_{\pm 0.7}$} & $93.4_{\pm 0.8}$ & $\mathbf{95.0_{\pm 0.6}}$ & $94.6_{\pm 0.2}$ & \underline{$96.0_{\pm 0.1}$} & $94.9_{\pm 0.2}$ & $\mathbf{96.1_{\pm 0.2}}$\\
        \midrule
        \multicolumn{9}{c}{\textbf{2-shot}} \\
        \midrule
        bottle      & $98.7_{\pm 0.8}$ & $99.2_{\pm 1.2}$ & $98.9_{\pm 0.7}$ & $99.8_{\pm 0.4}$ & $97.9_{\pm 0.1}$ & $98.4_{\pm 0.1}$ & $98.0_{\pm 0.1}$ & $98.1_{\pm 0.1}$\\
        cable       & $91.4_{\pm 2.2}$ & $91.6_{\pm 2.0}$ & $90.0_{\pm 2.5}$ & $93.3_{\pm 1.7}$ & $94.3_{\pm 0.5}$ & $93.8_{\pm 0.7}$ & $94.3_{\pm 0.6}$ & $93.9_{\pm 0.5}$\\
        capsule     & $76.5_{\pm 11.8}$ & $85.2_{\pm 10.4}$ & $74.1_{\pm 11.3}$ & $86.6_{\pm 8.3}$ & $90.6_{\pm 0.6}$ & $97.6_{\pm 0.2}$ & $93.2_{\pm 0.7}$ & $98.1_{\pm 0.2}$\\
        carpet      & $99.9_{\pm 0.1}$ & $99.0_{\pm 1.1}$ & $100.0_{\pm 0.0}$ & $99.6_{\pm 0.4}$ & $99.3_{\pm 0.1}$ & $98.7_{\pm 0.0}$ & $99.2_{\pm 0.0}$ & $98.7_{\pm 0.1}$\\
        grid        & $99.1_{\pm 0.6}$ & $99.4_{\pm 0.5}$ & $99.4_{\pm 0.4}$ & $98.0_{\pm 0.5}$ & $98.1_{\pm 0.2}$ & $99.0_{\pm 0.1}$ & $98.2_{\pm 0.1}$ & $98.9_{\pm 0.2}$\\
        hazelnut    & $99.3_{\pm 0.4}$ & $97.6_{\pm 2.0}$ & $99.7_{\pm 0.2}$ & $98.7_{\pm 1.3}$ & $98.8_{\pm 0.1}$ & $99.5_{\pm 0.1}$ & $98.5_{\pm 0.1}$ & $99.4_{\pm 0.0}$\\
        leather     & $100.0_{\pm 0.0}$ & $99.9_{\pm 0.2}$ & $100.0_{\pm 0.0}$ & $100.0_{\pm 0.0}$ & $99.3_{\pm 0.0}$ & $98.7_{\pm 0.1}$ & $99.4_{\pm 0.0}$ & $98.9_{\pm 0.1}$\\
        metal\_nut  & $100.0_{\pm 0.1}$ & $98.8_{\pm 0.6}$ & $99.0_{\pm 0.8}$ & $99.9_{\pm 0.2}$ & $88.5_{\pm 1.5}$ & $96.4_{\pm 0.7}$ & $91.6_{\pm 1.2}$ & $96.0_{\pm 0.8}$\\
        pill        & $96.4_{\pm 0.4}$ & $91.6_{\pm 3.7}$ & $93.7_{\pm 0.6}$ & $94.0_{\pm 2.6}$ & $94.2_{\pm 0.4}$ & $94.4_{\pm 0.8}$ & $93.3_{\pm 0.5}$ & $94.7_{\pm 0.8}$\\
        screw       & $67.3_{\pm 6.4}$ & $79.4_{\pm 3.7}$ & $79.5_{\pm 1.5}$ & $84.3_{\pm 4.7}$ & $92.9_{\pm 1.2}$ & $97.5_{\pm 0.3}$ & $95.2_{\pm 0.6}$ & $98.0_{\pm 0.2}$\\
        tile        & $99.6_{\pm 0.3}$ & $99.9_{\pm 0.1}$ & $98.7_{\pm 0.7}$ & $100.0_{\pm 0.0}$ & $95.5_{\pm 0.1}$ & $94.5_{\pm 0.2}$ & $95.4_{\pm 0.1}$ & $95.1_{\pm 0.1}$\\
        toothbrush  & $96.2_{\pm 1.3}$ & $96.0_{\pm 3.7}$ & $95.5_{\pm 1.0}$ & $96.8_{\pm 3.3}$ & $97.9_{\pm 0.3}$ & $99.0_{\pm 0.3}$ & $98.3_{\pm 0.3}$ & $98.4_{\pm 0.4}$\\
        transistor  & $75.6_{\pm 7.0}$ & $77.8_{\pm 8.0}$ & $80.7_{\pm 5.3}$ & $83.0_{\pm 7.8}$ & $86.6_{\pm 1.8}$ & $86.2_{\pm 2.2}$ & $84.4_{\pm 2.0}$ & $85.3_{\pm 2.3}$\\
        wood        & $98.6_{\pm 0.5}$ & $99.5_{\pm 0.3}$ & $99.0_{\pm 0.4}$ & $100.0_{\pm 0.0}$ & $95.3_{\pm 0.2}$ & $92.4_{\pm 0.8}$ & $95.0_{\pm 0.2}$ & $93.7_{\pm 0.9}$\\
        zipper      & $92.7_{\pm 4.0}$ & $99.4_{\pm 0.3}$ & $90.7_{\pm 6.4}$ & $99.7_{\pm 0.1}$ & $90.6_{\pm 0.7}$ & $96.7_{\pm 0.4}$ & $92.0_{\pm 0.5}$ & $97.0_{\pm 0.3}$\\
        \midrule
        \textbf{Average} & $92.8_{\pm 0.5}$ & \underline{$95.0_{\pm 0.0}$} & $93.1_{\pm 0.4}$ & $\mathbf{96.2_{\pm 0.4}}$ & $94.8_{\pm 0.1}$ & \underline{$96.3_{\pm 0.1}$} & $95.2_{\pm 0.2}$ & $\mathbf{96.4_{\pm 0.1}}$\\
        \midrule
        \multicolumn{9}{c}{\textbf{4-shot}} \\
        \midrule
        bottle      & $99.0_{\pm 0.8}$ & $99.5_{\pm 1.1}$ & $99.1_{\pm 0.7}$ & $99.9_{\pm 0.3}$ & $97.9_{\pm 0.1}$ & $98.5_{\pm 0.1}$ & $98.0_{\pm 0.1}$ & $98.2_{\pm 0.1}$\\
        cable       & $92.0_{\pm 2.2}$ & $92.1_{\pm 1.9}$ & $90.7_{\pm 2.4}$ & $93.9_{\pm 1.7}$ & $94.5_{\pm 0.6}$ & $94.0_{\pm 0.7}$ & $94.5_{\pm 0.6}$ & $94.1_{\pm 0.6}$\\
        capsule     & $78.6_{\pm 12.5}$ & $84.2_{\pm 11.3}$ & $76.0_{\pm 12.2}$ & $86.1_{\pm 9.3}$ & $90.8_{\pm 0.8}$ & $97.7_{\pm 0.3}$ & $93.3_{\pm 0.8}$ & $98.2_{\pm 0.2}$\\
        carpet      & $99.9_{\pm 0.2}$ & $99.1_{\pm 1.0}$ & $100.0_{\pm 0.1}$ & $99.6_{\pm 0.4}$ & $99.3_{\pm 0.1}$ & $98.8_{\pm 0.1}$ & $99.3_{\pm 0.0}$ & $98.8_{\pm 0.1}$\\
        grid        & $99.0_{\pm 0.5}$ & $99.6_{\pm 0.5}$ & $99.4_{\pm 0.2}$ & $98.2_{\pm 0.5}$ & $98.2_{\pm 0.2}$ & $98.8_{\pm 0.3}$ & $98.3_{\pm 0.1}$ & $98.9_{\pm 0.2}$\\
        hazelnut    & $99.5_{\pm 0.5}$ & $98.2_{\pm 1.9}$ & $99.8_{\pm 0.2}$ & $99.1_{\pm 1.2}$ & $98.8_{\pm 0.1}$ & $99.5_{\pm 0.1}$ & $98.5_{\pm 0.1}$ & $99.4_{\pm 0.1}$\\
        leather     & $100.0_{\pm 0.0}$ & $99.9_{\pm 0.2}$ & $100.0_{\pm 0.0}$ & $100.0_{\pm 0.0}$ & $99.3_{\pm 0.1}$ & $99.5_{\pm 0.1}$ & $99.4_{\pm 0.0}$ & $98.9_{\pm 0.1}$\\
        metal\_nut  & $100.0_{\pm 0.0}$ & $99.0_{\pm 0.6}$ & $99.4_{\pm 0.8}$ & $99.9_{\pm 0.2}$ & $89.5_{\pm 2.0}$ & $96.8_{\pm 0.8}$ & $92.4_{\pm 1.5}$ & $96.5_{\pm 0.9}$\\
        pill        & $96.4_{\pm 0.4}$ & $92.5_{\pm 3.3}$ & $93.9_{\pm 0.8}$ & $94.7_{\pm 2.4}$ & $94.5_{\pm 0.6}$ & $94.7_{\pm 0.8}$ & $93.7_{\pm 0.7}$ & $95.0_{\pm 0.8}$\\
        screw       & $71.3_{\pm 7.7}$ & $81.5_{\pm 4.5}$ & $80.6_{\pm 2.4}$ & $86.2_{\pm 5.0}$ & $93.8_{\pm 1.6}$ & $97.8_{\pm 0.5}$ & $95.6_{\pm 0.7}$ & $98.2_{\pm 0.3}$\\
        tile        & $99.7_{\pm 0.3}$ & $99.9_{\pm 0.1}$ & $99.0_{\pm 0.8}$ & $100.0_{\pm 0.0}$ & $95.5_{\pm 0.2}$ & $94.5_{\pm 0.2}$ & $95.5_{\pm 0.2}$ & $95.1_{\pm 0.2}$\\
        toothbrush  & $96.3_{\pm 1.1}$ & $96.2_{\pm 3.4}$ & $95.7_{\pm 1.2}$ & $97.0_{\pm 3.0}$ & $97.9_{\pm 0.3}$ & $99.0_{\pm 0.2}$ & $98.3_{\pm 0.2}$ & $98.4_{\pm 0.3}$\\
        transistor  & $80.0_{\pm 8.5}$ & $82.1_{\pm 9.1}$ & $83.7_{\pm 6.2}$ & $86.4_{\pm 8.0}$ & $87.9_{\pm 2.4}$ & $87.4_{\pm 2.5}$ & $85.9_{\pm 2.7}$ & $86.6_{\pm 2.7}$\\
        wood        & $98.3_{\pm 0.8}$ & $99.6_{\pm 0.3}$ & $98.9_{\pm 0.3}$ & $100.0_{\pm 0.0}$ & $95.3_{\pm 0.2}$ & $92.3_{\pm 0.7}$ & $95.1_{\pm 0.2}$ & $93.6_{\pm 0.8}$\\
        zipper      & $93.6_{\pm 3.5}$ & $99.5_{\pm 0.3}$ & $92.3_{\pm 5.7}$ & $99.8_{\pm 0.1}$ & $90.9_{\pm 0.9}$ & $97.0_{\pm 0.5}$ & $92.3_{\pm 0.7}$ & $97.3_{\pm 0.4}$\\
        \midrule
        \textbf{Average} & $95.2_{\pm 0.9}$ & \underline{$96.0_{\pm 1.2}$} & $95.2_{\pm 0.8}$ & $\mathbf{97.0_{\pm 0.8}}$ & $95.6_{\pm 0.1}$ & \underline{$96.8_{\pm 0.1}$} & $95.9_{\pm 0.1}$ & $\mathbf{96.9_{\pm 0.1}}$\\
        \bottomrule
    \end{tabular}
    }
    \end{center}
    \caption{Comparison of image-level and pixel-level AUC for each category in MVTec using training-free few-shot methods.}
    \label{tab:mvtec_few_shot}
\end{table*}
\clearpage

\begin{table*}[ht]
    \begin{center}
    \resizebox{0.8\linewidth}{!}{
    \setlength{\tabcolsep}{1.0mm}
    \begin{tabular}{ccccc|cccc}
        \toprule
        \multirow{3}{*}{Category} & \multicolumn{4}{c}{\textbf{Image-Level AUC}} & \multicolumn{4}{c}{\textbf{Pixel-Level AUC}}\\
        \cmidrule(lr){2-5} \cmidrule(lr){6-9}
         & PatchEAD & PatchEAD & PatchEAD+ & PatchEAD+ & PatchEAD & PatchEAD & PatchEAD+ & PatchEAD+ \\
         \cmidrule(lr){2-5} \cmidrule(lr){6-9}
         & ViT-B/16+ & DINOv2-B & ViT-B/16+ & DINOv2-B & ViT-B/16+ & DINOv2-B & ViT-B/16+ & DINOv2-B \\
        \midrule
        \multicolumn{9}{c}{\textbf{1-shot}} \\
        \midrule
        candle       & $92.0_{\pm 1.6}$ & $86.2_{\pm 3.6}$ & $94.8_{\pm 1.0}$ & $90.7_{\pm 1.1}$ & $96.8_{\pm 0.0}$ & $98.7_{\pm 0.1}$ & $97.0_{\pm 0.1}$ & $98.7_{\pm 0.1}$\\
        capsules     & $79.2_{\pm 1.1}$ & $95.7_{\pm 0.6}$ & $80.5_{\pm 2.4}$ & $93.1_{\pm 0.3}$ & $94.7_{\pm 0.3}$ & $97.3_{\pm 0.1}$ & $95.4_{\pm 0.1}$ & $96.7_{\pm 0.1}$\\
        cashew       & $92.4_{\pm 1.5}$ & $86.2_{\pm 1.8}$ & $93.4_{\pm 1.4}$ & $90.5_{\pm 1.1}$ & $97.0_{\pm 0.3}$ & $99.0_{\pm 0.1}$ & $97.5_{\pm 0.1}$ & $98.4_{\pm 0.1}$\\
        chewinggum   & $97.4_{\pm 0.0}$ & $98.2_{\pm 0.7}$ & $97.9_{\pm 0.1}$ & $98.7_{\pm 0.1}$ & $98.7_{\pm 0.1}$ & $99.4_{\pm 0.0}$ & $98.9_{\pm 0.1}$ & $99.4_{\pm 0.0}$ \\
        fryum        & $93.7_{\pm 2.3}$ & $94.2_{\pm 1.0}$ & $92.6_{\pm 1.5}$ & $95.7_{\pm 1.0}$ & $91.0_{\pm 0.7}$ & $92.7_{\pm 0.2}$ & $93.1_{\pm 0.4}$ & $93.8_{\pm 0.3}$\\
        macaroni1    & $85.1_{\pm 3.1}$ & $88.6_{\pm 2.2}$ & $91.1_{\pm 0.8}$ & $92.3_{\pm 1.1}$ & $96.7_{\pm 0.4}$ & $99.2_{\pm 0.0}$ & $97.6_{\pm 0.3}$ & $99.3_{\pm 0.0}$\\
        macaroni2    & $63.5_{\pm 4.1}$ & $62.9_{\pm 3.9}$ & $73.3_{\pm 2.4}$ & $67.4_{\pm 2.9}$ & $91.2_{\pm 0.9}$ & $96.9_{\pm 0.2}$ & $94.6_{\pm 0.1}$ & $98.3_{\pm 0.1}$\\
        pcb1         & $66.3_{\pm 17.8}$ & $75.6_{\pm 4.8}$ & $79.7_{\pm 8.6}$ & $82.1_{\pm 7.0}$ & $95.6_{\pm 1.1}$ & $99.0_{\pm 0.0}$ & $98.0_{\pm 0.3}$ & $99.0_{\pm 0.0}$ \\
        pcb2         & $74.9_{\pm 2.9}$ & $75.0_{\pm 3.5}$ & $72.7_{\pm 3.4}$ & $82.3_{\pm 3.7}$ & $92.2_{\pm 0.3}$ & $95.7_{\pm 0.4}$ & $92.8_{\pm 0.3}$ & $95.9_{\pm 0.4}$ \\
        pcb3         & $79.7_{\pm 4.6}$ & $77.5_{\pm 0.8}$ & $73.5_{\pm 5.6}$ & $86.8_{\pm 1.5}$ & $93.3_{\pm 0.6}$ & $96.7_{\pm 0.2}$ & $93.4_{\pm 0.7}$ & $96.2_{\pm 0.2}$\\
        pcb4         & $90.6_{\pm 4.3}$ & $87.2_{\pm 1.4}$ & $85.5_{\pm 4.3}$ & $93.1_{\pm 1.3}$ & $89.8_{\pm 1.0}$ & $94.4_{\pm 0.6}$ & $92.1_{\pm 0.8}$ & $94.7_{\pm 0.6}$\\
        pipe\_fryum  & $99.4_{\pm 0.2}$ & $97.6_{\pm 0.9}$ & $99.6_{\pm 0.2}$ & $95.8_{\pm 0.7}$ & $97.0_{\pm 0.4}$ & $98.7_{\pm 0.1}$ & $96.6_{\pm 0.6}$ & $97.9_{\pm 0.2}$\\
        \midrule
        \textbf{Average} & $84.5_{\pm 1.3}$ & $85.4_{\pm 0.3}$ & $\underline{86.2_{\pm 0.6}}$ & $\mathbf{89.1_{\pm 0.4}}$ & $94.5_{\pm 0.2}$ & $\underline{97.3_{\pm 0.1}}$ & $95.6_{\pm 0.1}$ & $\mathbf{97.4_{\pm 0.1}}$\\
        \midrule
        \multicolumn{9}{c}{\textbf{2-shot}} \\
        \midrule
        candle       & $92.2_{\pm 1.4}$ & $88.9_{\pm 3.7}$ & $94.6_{\pm 1.1}$ & $91.7_{\pm 1.4}$ & $97.0_{\pm 0.2}$ & $98.8_{\pm 0.2}$ & $97.2_{\pm 0.2}$ & $98.8_{\pm 0.2}$\\
        capsules     & $79.8_{\pm 1.8}$ & $96.0_{\pm 1.3}$ & $80.7_{\pm 2.6}$ & $93.1_{\pm 0.6}$ & $94.7_{\pm 0.3}$ & $97.3_{\pm 0.3}$ & $95.4_{\pm 0.2}$ & $96.7_{\pm 0.3}$\\
        cashew       & $92.7_{\pm 1.7}$ & $87.8_{\pm 2.8}$ & $94.6_{\pm 1.7}$ & $91.6_{\pm 1.7}$ & $97.2_{\pm 0.3}$ & $99.1_{\pm 0.1}$ & $97.6_{\pm 0.2}$ & $98.6_{\pm 0.2}$\\
        chewinggum   & $97.1_{\pm 0.5}$ & $97.9_{\pm 0.7}$ & $97.5_{\pm 0.5}$ & $98.7_{\pm 0.2}$ & $98.7_{\pm 0.1}$ & $99.4_{\pm 0.1}$ & $98.8_{\pm 0.1}$ & $99.3_{\pm 0.0}$ \\
        fryum        & $94.7_{\pm 1.9}$ & $95.5_{\pm 1.5}$ & $93.0_{\pm 1.2}$ & $96.8_{\pm 1.3}$ & $91.6_{\pm 0.8}$ & $93.4_{\pm 0.7}$ & $93.5_{\pm 0.5}$ & $94.3_{\pm 0.6}$\\
        macaroni1    & $87.1_{\pm 3.6}$ & $88.9_{\pm 1.9}$ & $90.4_{\pm 1.7}$ & $92.9_{\pm 1.2}$ & $97.1_{\pm 0.8}$ & $99.3_{\pm 0.1}$ & $97.8_{\pm 0.6}$ & $99.4_{\pm 0.1}$\\
        macaroni2    & $71.6_{\pm 8.7}$ & $66.5_{\pm 4.9}$ & $76.8_{\pm 3.9}$ & $69.6_{\pm 3.2}$ & $92.9_{\pm 1.9}$ & $97.3_{\pm 0.5}$ & $95.3_{\pm 0.8}$ & $98.6_{\pm 0.3}$\\
        pcb1         & $73.8_{\pm 15.0}$ & $76.7_{\pm 3.6}$ & $81.7_{\pm 6.9}$ & $83.5_{\pm 5.4}$ & $96.8_{\pm 1.4}$ & $99.1_{\pm 0.1}$ & $98.3_{\pm 0.4}$ & $99.1_{\pm 0.1}$ \\
        pcb2         & $74.5_{\pm 2.7}$ & $75.7_{\pm 4.1}$ & $71.4_{\pm 3.3}$ & $83.2_{\pm 3.2}$ & $92.9_{\pm 0.8}$ & $96.2_{\pm 0.6}$ & $93.3_{\pm 0.7}$ & $96.4_{\pm 0.6}$ \\
        pcb3         & $83.3_{\pm 5.0}$ & $80.1_{\pm 3.5}$ & $77.1_{\pm 5.5}$ & $87.5_{\pm 1.9}$ & $93.7_{\pm 0.6}$ & $96.8_{\pm 0.2}$ & $93.9_{\pm 0.7}$ & $96.3_{\pm 0.2}$\\
        pcb4         & $87.4_{\pm 12.4}$ & $87.3_{\pm 1.2}$ & $85.2_{\pm 11.1}$ & $93.3_{\pm 1.2}$ & $90.9_{\pm 1.4}$ & $94.5_{\pm 0.5}$ & $92.9_{\pm 1.1}$ & $94.9_{\pm 0.5}$\\
        pipe\_fryum  & $99.5_{\pm 0.2}$ & $97.5_{\pm 0.7}$ & $99.7_{\pm 0.2}$ & $95.9_{\pm 0.5}$ & $97.2_{\pm 0.5}$ & $98.7_{\pm 0.1}$ & $96.8_{\pm 0.6}$ & $98.0_{\pm 0.2}$\\
        \midrule
        \textbf{Average} & $\underline{87.8_{\pm 1.4}}$ & $87.7_{\pm 0.3}$ & $87.6_{\pm 1.2}$ & $\mathbf{90.6_{\pm 0.5}}$ & $95.6_{\pm 0.1}$ & $\mathbf{97.7_{\pm 0.1}}$ & $\underline{96.2_{\pm 0.1}}$ & $\mathbf{97.7_{\pm 0.1}}$\\
        \midrule
        \multicolumn{9}{c}{\textbf{4-shot}} \\
        \midrule
        candle       & $92.9_{\pm 1.7}$ & $90.2_{\pm 3.6}$ & $94.9_{\pm 1.2}$ & $92.3_{\pm 1.5}$ & $97.2_{\pm 0.4}$ & $98.9_{\pm 0.2}$ & $97.4_{\pm 0.4}$ & $98.9_{\pm 0.2}$\\
        capsules     & $80.5_{\pm 2.2}$ & $96.2_{\pm 1.1}$ & $81.6_{\pm 2.5}$ & $93.4_{\pm 0.7}$ & $95.0_{\pm 0.6}$ & $97.4_{\pm 0.3}$ & $95.7_{\pm 0.5}$ & $96.8_{\pm 0.3}$\\
        cashew       & $92.8_{\pm 1.9}$ & $88.5_{\pm 2.6}$ & $94.7_{\pm 1.6}$ & $92.3_{\pm 1.7}$ & $97.2_{\pm 0.3}$ & $99.1_{\pm 0.2}$ & $97.6_{\pm 0.2}$ & $98.6_{\pm 0.2}$\\
        chewinggum   & $97.5_{\pm 0.7}$ & $98.2_{\pm 0.7}$ & $97.6_{\pm 0.5}$ & $98.7_{\pm 0.3}$ & $98.7_{\pm 0.1}$ & $99.4_{\pm 0.0}$ & $98.8_{\pm 0.1}$ & $99.3_{\pm 0.0}$ \\
        fryum        & $95.1_{\pm 1.7}$ & $95.9_{\pm 1.5}$ & $93.6_{\pm 1.4}$ & $97.0_{\pm 1.1}$ & $92.2_{\pm 1.0}$ & $93.8_{\pm 0.9}$ & $94.0_{\pm 0.8}$ & $94.7_{\pm 0.7}$\\
        macaroni1    & $87.3_{\pm 3.0}$ & $89.1_{\pm 1.7}$ & $90.3_{\pm 1.4}$ & $93.2_{\pm 1.1}$ & $97.2_{\pm 0.7}$ & $99.4_{\pm 0.1}$ & $97.9_{\pm 0.5}$ & $99.4_{\pm 0.1}$\\
        macaroni2    & $73.0_{\pm 7.5}$ & $67.6_{\pm 4.4}$ & $77.0_{\pm 3.2}$ & $70.8_{\pm 3.8}$ & $93.6_{\pm 1.9}$ & $97.5_{\pm 0.5}$ & $95.4_{\pm 0.7}$ & $98.7_{\pm 0.3}$\\
        pcb1         & $78.8_{\pm 14.2}$ & $79.1_{\pm 4.8}$ & $84.4_{\pm 7.0}$ & $85.9_{\pm 5.8}$ & $97.1_{\pm 1.3}$ & $99.2_{\pm 0.1}$ & $98.5_{\pm 0.4}$ & $99.2_{\pm 0.1}$ \\
        pcb2         & $76.6_{\pm 4.1}$ & $77.9_{\pm 4.7}$ & $74.5_{\pm 5.3}$ & $84.7_{\pm 3.4}$ & $93.5_{\pm 1.1}$ & $96.5_{\pm 0.7}$ & $93.9_{\pm 1.0}$ & $96.7_{\pm 0.7}$ \\
        pcb3         & $84.7_{\pm 4.6}$ & $80.4_{\pm 2.9}$ & $80.0_{\pm 6.1}$ & $88.2_{\pm 1.9}$ & $94.1_{\pm 0.7}$ & $97.1_{\pm 0.4}$ & $94.4_{\pm 0.8}$ & $96.5_{\pm 0.4}$\\
        pcb4         & $87.3_{\pm 10.3}$ & $87.6_{\pm 1.1}$ & $84.5_{\pm 9.2}$ & $93.6_{\pm 1.1}$ & $91.8_{\pm 1.8}$ & $94.9_{\pm 0.6}$ & $93.6_{\pm 1.3}$ & $95.2_{\pm 0.6}$\\
        pipe\_fryum  & $99.6_{\pm 0.2}$ & $97.5_{\pm 1.0}$ & $99.8_{\pm 0.2}$ & $96.2_{\pm 0.9}$ & $97.3_{\pm 0.4}$ & $98.8_{\pm 0.1}$ & $96.9_{\pm 0.5}$ & $98.1_{\pm 0.2}$\\
        \midrule
        \textbf{Average} & $89.3_{\pm 0.5}$ & $88.9_{\pm 0.0}$ & $\underline{89.5_{\pm 0.3}}$ & $\mathbf{91.9_{\pm 0.3}}$ & $96.1_{\pm 0.1}$ & $\mathbf{98.0_{\pm 0.0}}$ & $\underline{96.7_{\pm 0.1}}$ & $\mathbf{98.0_{\pm 0.0}}$\\
        \bottomrule
    \end{tabular}
    }
    \end{center}
    \caption{Comparison of image-level and pixel-level AUC for each category in VisA using training-free few-shot methods.}
    \label{tab:visa_few_shot}
\end{table*}
\clearpage

\begin{table*}[ht]
    \begin{center}
    \resizebox{0.85\linewidth}{!}{
    \setlength{\tabcolsep}{1mm}
    \begin{tabular}{ccccc|cccc}
        \toprule
        \multirow{3}{*}{Category} & \multicolumn{4}{c}{\textbf{Image-Level AUC}} & \multicolumn{4}{c}{\textbf{Pixel-Level AUC}}\\
        \cmidrule(lr){2-5} \cmidrule(lr){6-9}
        & PatchEAD & PatchEAD & PatchEAD+ & PatchEAD+ & PatchEAD & PatchEAD & PatchEAD+ & PatchEAD+ \\
        \cmidrule(lr){2-5} \cmidrule(lr){6-9}
        & ViT-B/16+ & DINOv2-B & ViT-B/16+ & DINOv2-B & ViT-B/16+ & DINOv2-B & ViT-B/16+ & DINOv2-B \\
        \midrule
        \multicolumn{9}{c}{\textbf{MVTec}} \\
        \midrule
        bottle       & $98.7$ & $99.5$ & $97.7$ & $99.8$ & $98.2$ & $98.7$ & $98.2$ & $97.9$\\
        cable        & $93.5$ & $96.4$ & $92.4$ & $95.1$ & $95.2$ & $95.3$ & $94.3$ & $93.0$\\
        capsule      & $88.0$ & $86.9$ & $91.1$ & $89.1$ & $92.8$ & $97.7$ & $92.1$ & $98.5$\\
        carpet       & $98.2$ & $98.6$ & $99.1$ & $99.9$ & $98.9$ & $98.3$ & $99.0$ & $98.6$\\
        grid         & $98.5$ & $90.1$ & $93.9$ & $99.7$ & $97.9$ & $98.6$ & $97.8$ & $99.0$\\
        hazelnut     & $98.2$ & $90.2$ & $98.2$ & $95.1$ & $99.1$ & $99.4$ & $99.1$ & $99.3$\\
        leather      & $99.4$ & $95.1$ & $96.9$ & $100.0$ & $98.9$ & $98.5$ & $99.0$ & $99.3$\\
        metal\_nut   & $95.4$ & $96.2$ & $97.1$ & $95.4$ & $79.4$ & $87.6$ & $78.4$ & $85.4$\\
        pill         & $95.2$ & $95.2$ & $95.2$ & $95.6$ & $95.0$ & $93.0$ & $94.9$ & $95.2$\\
        screw        & $80.9$ & $81.1$ & $82.5$ & $87.0$ & $97.0$ & $97.0$ & $93.6$ & $97.3$\\
        tile         & $100.0$ & $100.0$ & $99.8$ & $99.8$ & $96.0$ & $96.7$ & $95.6$ & $95.9$\\
        toothbrush   & $100.0$ & $100.0$ & $99.2$ & $98.9$ & $98.4$ & $98.8$ & $98.4$ & $97.4$\\
        transistor   & $89.4$ & $91.9$ & $90.0$ & $92.7$ & $91.3$ & $92.5$ & $89.7$ & $88.9$\\
        wood         & $92.2$ & $93.9$ & $94.9$ & $97.3$ & $96.1$ & $92.8$ & $96.0$ & $91.0$\\
        zipper       & $90.3$ & $98.7$ & $91.3$ & $93.9$ & $90.1$ & $93.2$ & $90.1$ & $97.1$\\
        \midrule
        \textbf{Average} & $94.5$ & $94.3$ & $\underline{94.6}$ & $\mathbf{95.9}$ & $95.0$ & $\mathbf{95.9}$ & $94.4$ & $\underline{95.6}$\\
        \midrule
        \multicolumn{9}{c}{\textbf{VisA}} \\
        \midrule
        candle       & $93.1$ & $90.8$ & $93.3$ & $89.8$ & $97.7$ & $98.2$ & $97.7$ & $99.0$\\
        capsules     & $80.7$ & $92.5$ & $81.7$ & $94.6$ & $96.2$ & $96.0$ & $95.7$ & $97.0$\\
        cashew       & $86.0$ & $88.7$ & $84.3$ & $93.7$ & $97.9$ & $99.4$ & $97.7$ & $98.5$\\
        chewinggum   & $97.7$ & $97.7$ & $98.1$ & $98.8$ & $98.5$ & $98.8$ & $98.6$ & $99.0$\\
        fryum        & $95.6$ & $92.7$ & $92.3$ & $96.0$ & $94.1$ & $92.7$ & $93.1$ & $94.6$\\
        macaroni1    & $87.5$ & $85.3$ & $90.8$ & $96.7$ & $98.0$ & $98.5$ & $98.0$ & $99.5$\\
        macaroni2    & $75.3$ & $67.2$ & $76.2$ & $82.4$ & $95.6$ & $97.5$ & $95.6$ & $98.9$\\
        pcb1         & $81.3$ & $79.5$ & $82.9$ & $88.9$ & $98.2$ & $99.0$ & $98.4$ & $99.3$\\
        pcb2         & $81.4$ & $81.6$ & $81.7$ & $87.3$ & $95.0$ & $95.8$ & $95.0$ & $97.2$\\
        pcb3         & $88.2$ & $91.2$ & $90.5$ & $88.2$ & $95.5$ & $96.8$ & $95.2$ & $96.6$\\
        pcb4         & $78.3$ & $87.9$ & $75.0$ & $93.8$ & $93.3$ & $90.9$ & $92.6$ & $94.8$\\
        pipe\_fryum  & $96.8$ & $94.5$ & $96.4$ & $96.4$ & $97.9$ & $99.0$ & $97.8$ & $98.0$\\
        \midrule
        \textbf{Average} & $86.8$ & $\underline{87.5}$ & $86.9$ & $\mathbf{92.2}$ & $96.5$ & $\underline{96.9}$ & $96.3$ & $\mathbf{97.7}$\\
        \midrule
        \multicolumn{9}{c}{\textbf{MPDD}} \\
        \midrule
        bracket\_black & $50.9$ & $46.3$ & $53.4$ & $45.8$ & $89.8$ & $95.3$ & $91.4$ & $96.6$\\
        bracket\_brown & $50.5$ & $50.7$ & $51.7$ & $59.7$ & $93.8$ & $94.0$ & $93.8$ & $94.3$\\
        bracket\_white & $43.1$ & $35.7$ & $45.0$ & $43.2$ & $92.2$ & $97.6$ & $92.8$ & $98.6$\\
        connector      & $90.4$ & $77.7$ & $90.5$ & $79.5$ & $97.5$ & $97.7$ & $97.6$ & $97.7$\\
        metal\_plate   & $90.4$ & $88.6$ & $91.5$ & $88.4$ & $92.2$ & $91.9$ & $91.7$ & $93.1$\\
        tubes          & $91.7$ & $92.9$ & $91.7$ & $94.9$ & $98.5$ & $98.8$ & $98.6$ & $99.2$\\
        \midrule
        \textbf{Average} & $\underline{69.5}$ & $65.3$ & $\mathbf{70.6}$ & $68.6$ & $94.0$ & $\underline{95.9}$ & $94.3$ & $\mathbf{96.6}$\\
        \bottomrule
    \end{tabular}
    }
    \end{center}
    \caption{Comparison of image-level and pixel-level AUC for each category in MVTec, VisA, and MPDD using training-free zero-shot methods.}
    \label{tab:mvtec_visa_mpdd_zero_shot}
\end{table*}
\clearpage

\begin{table*}[ht]
    \begin{center}
    \resizebox{0.85\linewidth}{!}{
    \setlength{\tabcolsep}{1mm}
    \begin{tabular}{ccccc|cccc}
        \toprule
        \multirow{3}{*}{Category} & \multicolumn{4}{c}{\textbf{Image-Level AUC}} & \multicolumn{4}{c}{\textbf{Pixel-Level AUC}}\\
        \cmidrule(lr){2-5} \cmidrule(lr){6-9}
         & PatchEAD & PatchEAD & PatchEAD+ & PatchEAD+ & PatchEAD & PatchEAD & PatchEAD+ & PatchEAD+ \\
        \cmidrule(lr){2-5} \cmidrule(lr){6-9}
         & ViT-B/16+ & DINOv2-B & ViT-B/16+ & DINOv2-B & ViT-B/16+ & DINOv2-B & ViT-B/16+ & DINOv2-B \\
        \midrule
        \multicolumn{9}{c}{\textbf{BTAD}} \\
        \midrule
        01 & $96.4$ & $95.7$ & $93.5$ & $94.1$ & $95.3$ & $97.3$ & $95.3$ & $97.0$\\
        02 & $81.0$ & $82.2$ & $81.7$ & $81.4$ & $94.6$ & $95.3$ & $95.0$ & $94.4$\\
        03 & $99.5$ & $99.2$ & $99.2$ & $98.1$ & $99.4$ & $99.6$ & $99.4$ & $99.5$\\
        \midrule
        \textbf{Average} & $\mathbf{92.3}$ & $\mathbf{92.3}$ & $91.5$ & $91.2$ & $96.4$ & $\mathbf{97.4}$ & $96.5$ & $\underline{97.0}$\\
        \midrule
        \multicolumn{9}{c}{\textbf{KSDD}} \\
        \midrule
        KSDD & $89.0$ & $93.7$ & $89.4$ & $97.0$ & $97.6$ & $\underline{99.2}$ & $97.7$ & $\mathbf{99.3}$\\
        \midrule
        \textbf{Average} & $89.0$ & $\underline{93.7}$ & $89.4$ & $\mathbf{97.0}$ & $97.6$ & $\underline{99.2}$ & $97.7$ & $\mathbf{99.3}$\\
        \midrule
        \multicolumn{9}{c}{\textbf{DAGM}} \\
        \midrule
        Class1  & $78.2$ & $85.1$ & $83.6$ & $88.3$ & $86.7$ & $80.8$ & $87.5$ & $76.2$\\
        Class2  & $99.4$ & $100.0$ & $99.6$ & $99.8$ & $99.2$ & $99.8$ & $99.2$ & $99.6$\\
        Class3  & $97.1$ & $99.6$ & $95.3$ & $99.9$ & $95.3$ & $97.8$ & $95.7$ & $97.3$\\
        Class4  & $100.0$ & $100.0$ & $100.0$ & $100.0$ & $99.2$ & $97.1$ & $99.0$ & $95.8$\\
        Class5  & $97.7$ & $99.9$ & $98.8$ & $100.0$ & $96.0$ & $99.8$ & $96.1$ & $99.7$\\
        Class6  & $91.6$ & $100.0$ & $95.3$ & $100.0$ & $91.6$ & $99.6$ & $92.1$ & $99.5$\\
        Class7  & $99.9$ & $100.0$ & $100.0$ & $100.0$ & $95.9$ & $97.8$ & $95.7$ & $96.8$\\
        Class8  & $82.4$ & $99.8$ & $83.7$ & $99.0$ & $93.1$ & $99.7$ & $93.5$ & $99.7$\\
        Class9  & $78.5$ & $91.6$ & $87.3$ & $95.9$ & $92.2$ & $99.6$ & $95.0$ & $99.9$\\
        Class10 & $100.0$ & $100.0$ & $99.8$ & $100.0$ & $99.6$ & $99.7$ & $99.6$ & $99.3$\\
        \midrule
        \textbf{Average} & $92.5$ & $\underline{97.6}$ & $94.3$ & $\mathbf{98.3}$ & $94.9$ & $\mathbf{97.2}$ & $95.3$ & $\underline{96.4}$\\
        \midrule
        \multicolumn{9}{c}{\textbf{DTD-Synthetic}} \\
        \midrule
        Blotchy\_099     & $93.7$ & $94.1$ & $97.6$ & $99.3$ & $99.0$ & $97.8$ & $99.0$ & $99.0$\\
        Fibrous\_183     & $97.6$ & $90.4$ & $98.1$ & $96.7$ & $98.7$ & $98.7$ & $98.6$ & $99.1$\\
        Marbled\_078     & $86.0$ & $88.6$ & $86.7$ & $97.4$ & $96.9$ & $96.2$ & $96.0$ & $97.9$\\
        Matted\_069      & $75.7$ & $73.1$ & $76.5$ & $92.2$ & $95.4$ & $94.2$ & $95.1$ & $98.7$\\
        Mesh\_114        & $88.2$ & $95.6$ & $87.4$ & $95.9$ & $97.7$ & $98.4$ & $97.1$ & $97.1$\\
        Perforated\_037  & $97.3$ & $94.0$ & $98.4$ & $98.3$ & $99.2$ & $99.0$ & $99.1$ & $98.7$\\
        Stratified\_154  & $90.6$ & $98.5$ & $84.9$ & $98.8$ & $99.2$ & $99.3$ & $99.2$ & $99.8$\\
        Woven\_001       & $98.9$ & $97.5$ & $98.5$ & $99.6$ & $99.4$ & $99.8$ & $99.2$ & $99.8$\\
        Woven\_068       & $94.4$ & $91.3$ & $94.5$ & $95.4$ & $98.7$ & $98.7$ & $98.3$ & $98.1$\\
        Woven\_104       & $96.5$ & $85.5$ & $90.9$ & $98.9$ & $97.2$ & $97.4$ & $97.6$ & $98.6$\\
        Woven\_125       & $94.9$ & $91.8$ & $94.2$ & $99.2$ & $99.0$ & $99.0$ & $98.8$ & $98.5$\\
        Woven\_127       & $95.1$ & $95.9$ & $94.7$ & $99.0$ & $94.2$ & $95.7$ & $94.3$ & $94.9$\\
        \midrule
        \textbf{Average} & $\underline{92.4}$ & $91.4$ & $91.9$ & $\mathbf{97.6}$ & $\underline{97.9}$ & $\underline{97.9}$ & $97.7$ & $\mathbf{98.3}$\\
        \bottomrule
    \end{tabular}
    }
    \end{center}
    \caption{Comparison of image-level and pixel-level AUC for each category in BTAD, KSDD, DAGM, and DTD-Synthetic using training-free zero-shot methods.}
    \label{tab:btad_ksdd_dagm_dtdsynthetic_zero_shot}
\end{table*}

\clearpage




\begin{figure*}[t]
    \begin{center}
      \includegraphics[width=\textwidth]{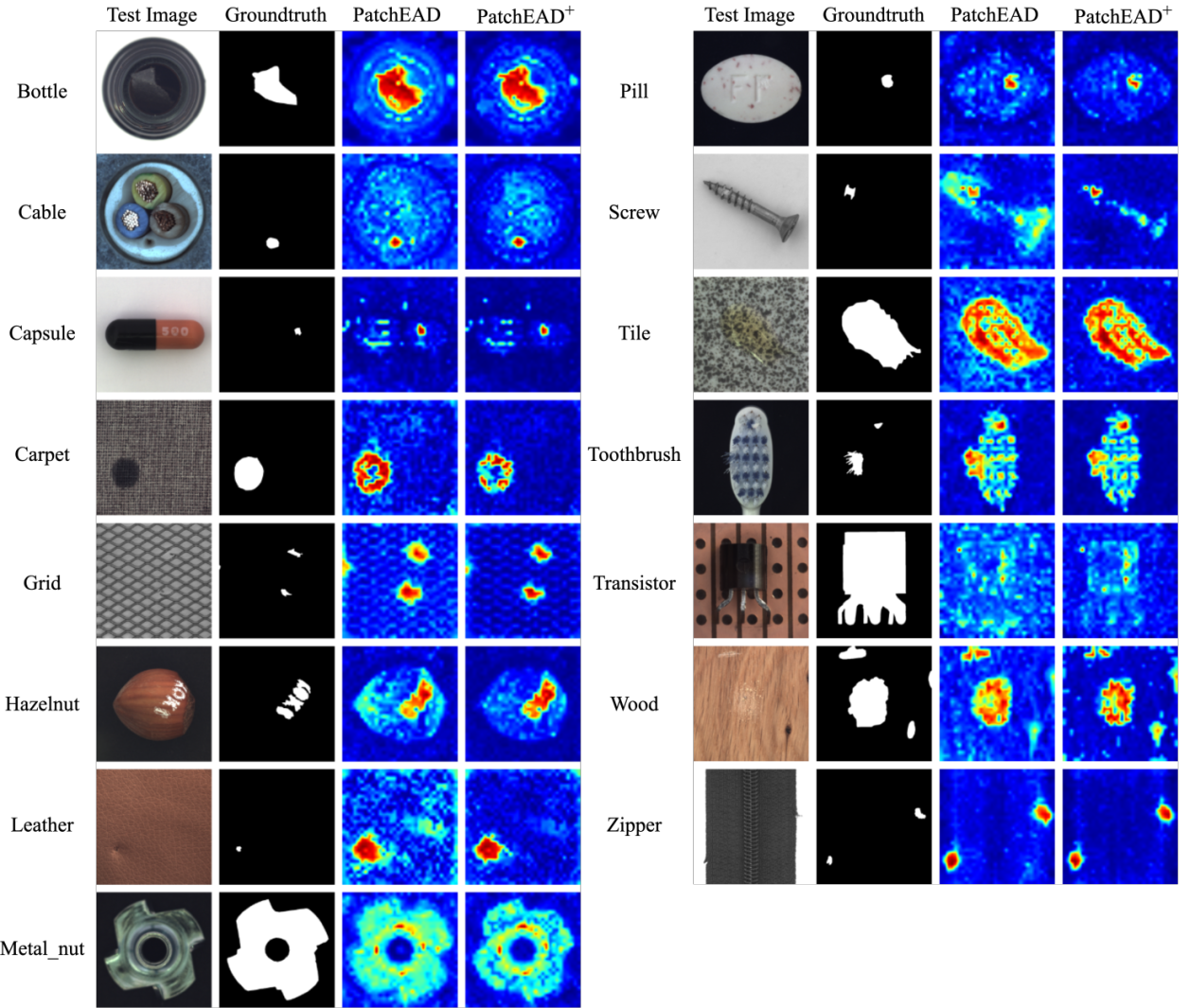}
    \end{center}
    \caption{
    Additional visualization results from PatchEAD and PatchEAD+(4-shot), tested on MVTec.
    }
    \label{fig:mvtec_few}
\end{figure*}
\clearpage

\begin{figure*}[t]
    \begin{center}
      \includegraphics[width=\textwidth]{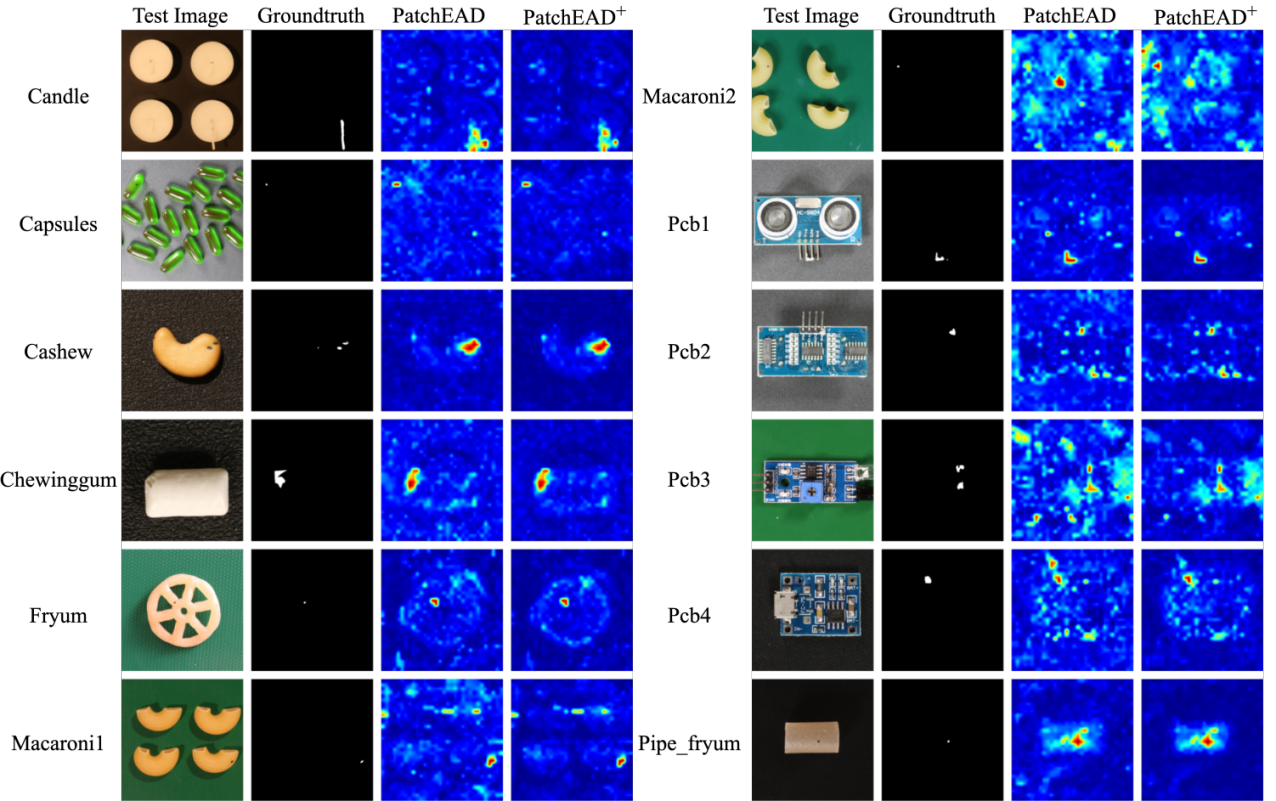}
    \end{center}
    \caption{
    Additional visualization results from PatchEAD and PatchEAD+(4-shot), tested on VisA.
    }
    \label{fig:visa_few}
\end{figure*}
\clearpage


\begin{figure*}[t]
    \begin{center}
      \includegraphics[width=\textwidth]{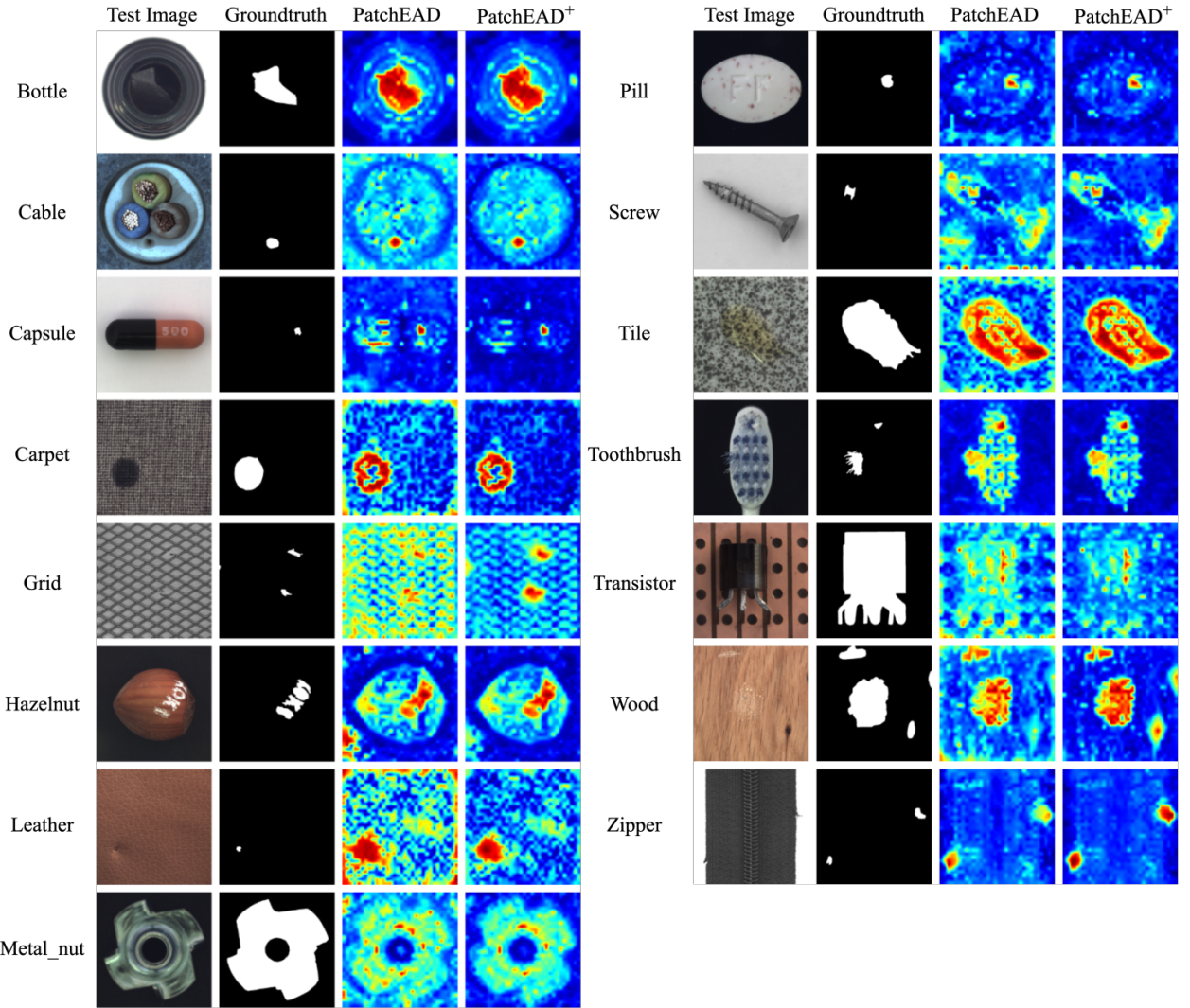}
    \end{center}
    \caption{
    Additional visualization results from PatchEAD and PatchEAD+(0-shot), tested on MVTec.
    }
    \label{fig:mvtec_zero}
\end{figure*}
\clearpage

\begin{figure*}[t]
    \begin{center}
      \includegraphics[width=\textwidth]{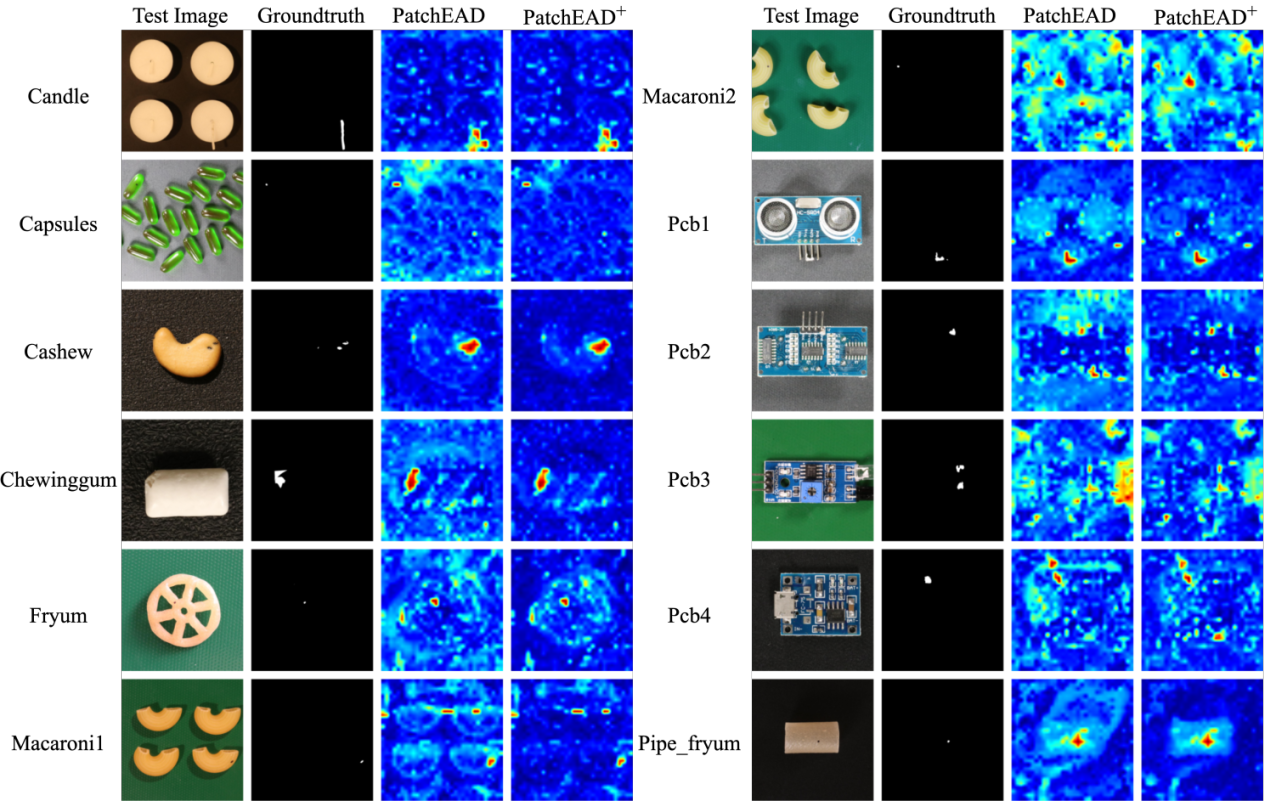}
    \end{center}
    \caption{
    Additional visualization results from PatchEAD and PatchEAD+(0-shot), tested on VisA.
    }
    \label{fig:visa_zero}
\end{figure*}

\begin{figure*}[t]
    \begin{center}
      \includegraphics[width=\textwidth]{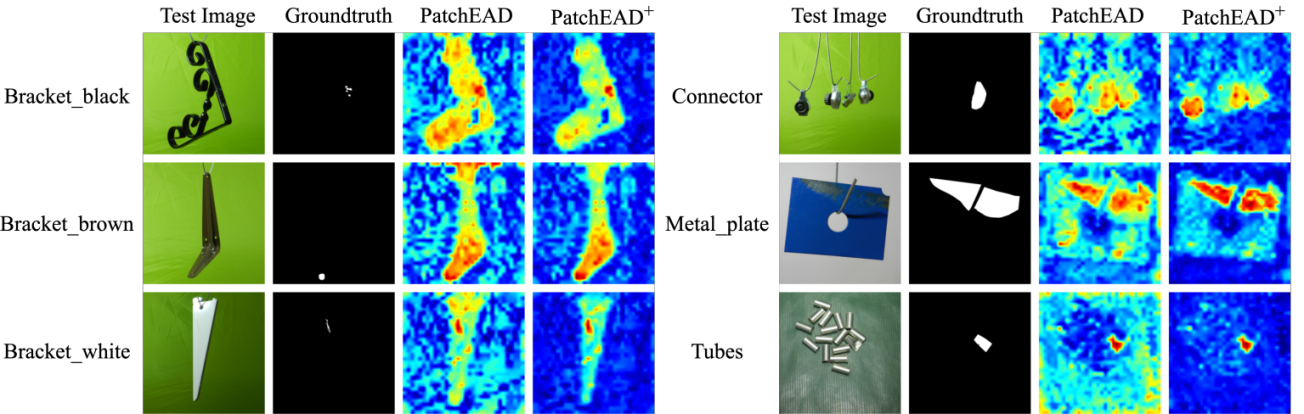}
    \end{center}
    \caption{
    Additional visualization results from PatchEAD and PatchEAD+(0-shot), tested on MPDD.
    }
    \label{fig:mpdd_zero}
\end{figure*}
\clearpage
    
\begin{figure*}[t]
    \begin{center}
      \includegraphics[width=0.9\textwidth]{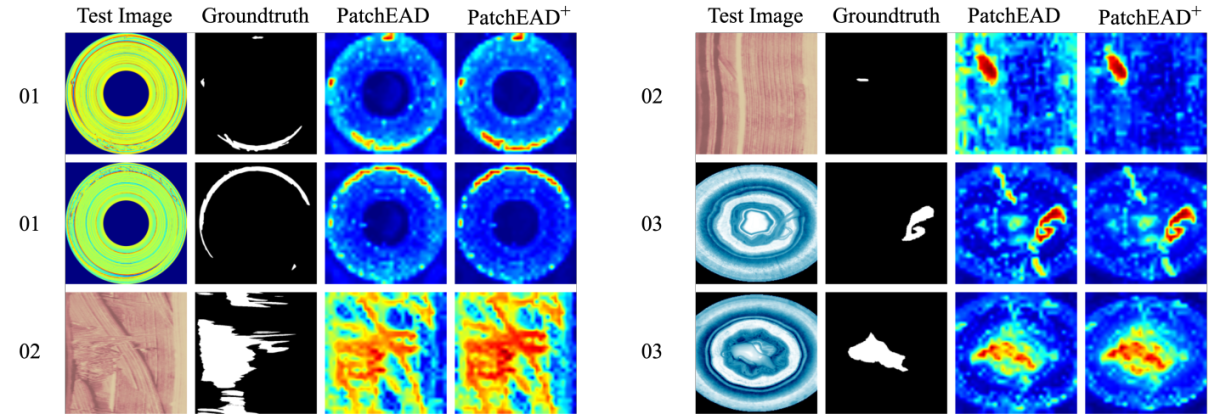}
    \end{center}
    \caption{
    Additional visualization results from PatchEAD and PatchEAD+(0-shot), tested on BTAD.
    }
    \label{fig:btad_zero}
\end{figure*}

\begin{figure*}[t]
    \begin{center}
      \includegraphics[width=0.9\textwidth]{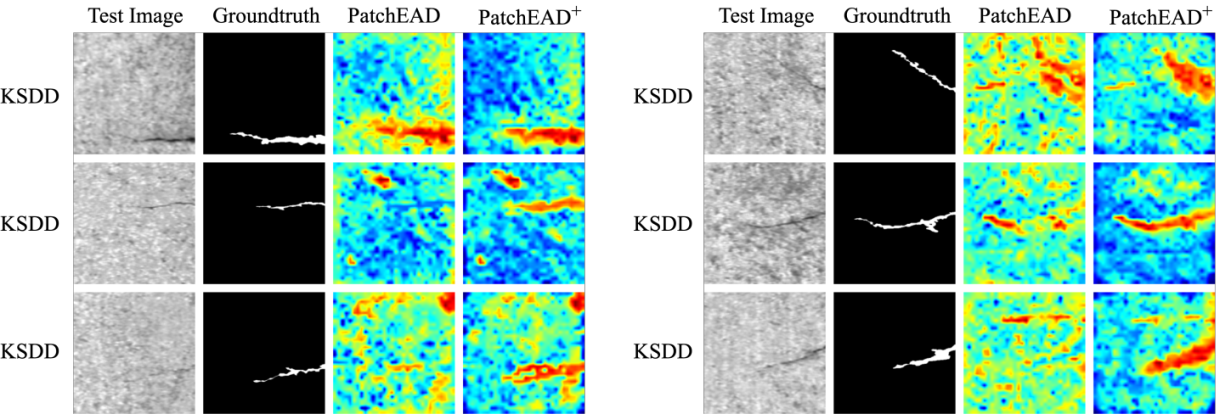}
    \end{center}
    \caption{
    Additional visualization results from PatchEAD and PatchEAD+(0-shot), tested on KSDD.
    }
    \label{fig:ksdd_zero}
\end{figure*}

\begin{figure*}[t]
    \begin{center}
      \includegraphics[width=0.9\textwidth]{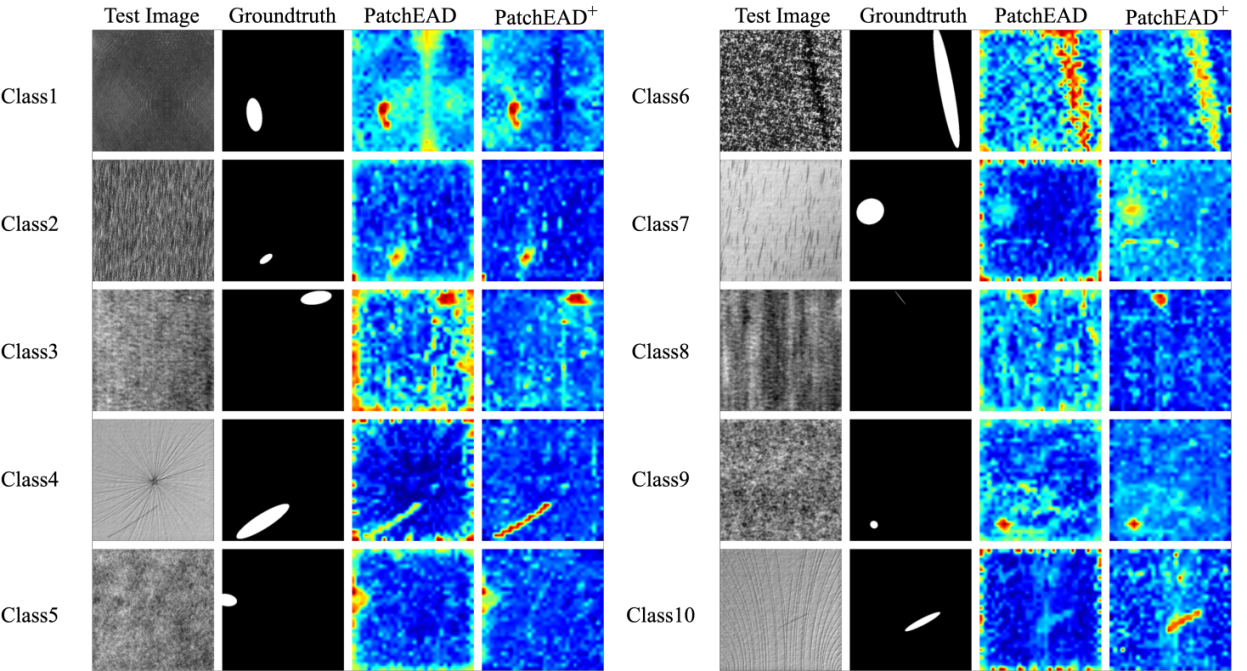}
    \end{center}
    \caption{
    Additional visualization results from PatchEAD and PatchEAD+(0-shot), tested on DAGM.
    }
    \label{fig:dagm_zero}
\end{figure*}

\clearpage

\begin{figure*}[t]
    \begin{center}
      \includegraphics[width=\textwidth]{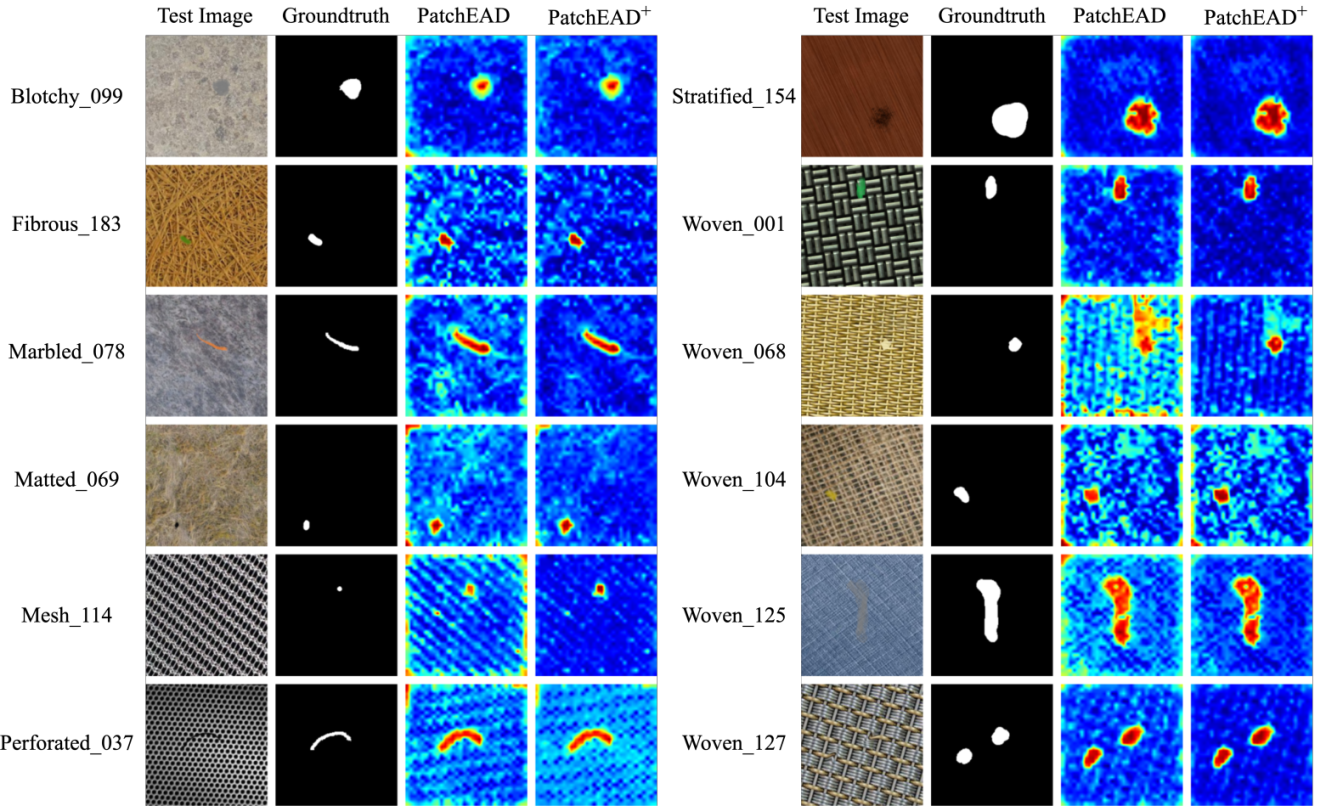}
    \end{center}
    \caption{
    Additional visualization results from PatchEAD and PatchEAD+(0-shot), tested on DTD-Synthetic.
    }
    \label{fig:dtd_synthetic_zero}
\end{figure*}

\begin{figure*}[t]
    \begin{center}
      \includegraphics[width=\textwidth]{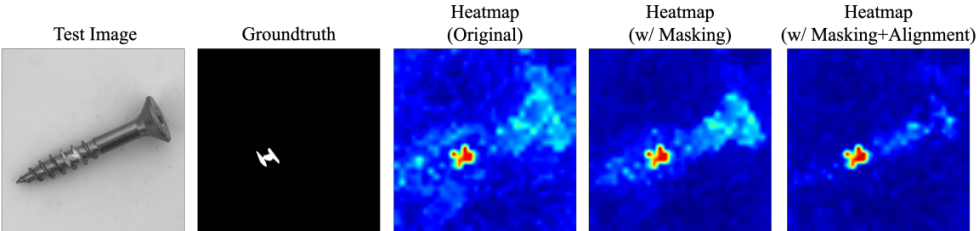}
    \end{center}
    \caption{
    Additional visualization results for the comparison to evaluate the effectiveness of Alignment and Masking.
    }
    \label{fig:case_study}
\end{figure*}
\clearpage


%

\end{document}